\documentclass{article} \usepackage{iclr2027_conference,times}
\iclrfinalcopy
\usepackage{amsmath,amsfonts,bm}

\def\eqref#1{equation~\ref{#1}}

\def\1{\bm{1}}

\DeclareMathAlphabet{\mathsfit}{\encodingdefault}{\sfdefault}{m}{sl}
\SetMathAlphabet{\mathsfit}{bold}{\encodingdefault}{\sfdefault}{bx}{n}

\usepackage{hyperref}
\usepackage{url}
\usepackage{xcolor}         
\usepackage{amsmath}
\usepackage{algorithm}
\usepackage{algorithmic}
\usepackage{wrapfig}
\usepackage{cleveref}
\usepackage{hhline}
\usepackage{graphicx}        
\usepackage{multirow}        
\usepackage{colortbl}
\usepackage{booktabs}      
\usepackage{pifont}
\usepackage{ dsfont }

\definecolor{myPurple}{RGB}{132,92,151}

\newcommand{\YSR}[1]{}
\newcommand{\RM}[1]{}
\newcommand{\LX}[1]{}
\newcommand{\PP}[1]{}

\definecolor{mygreen}{HTML}{66cc66}
\newcommand{\eg}{\textit{e.g.} }
\newcommand{\etc}{\textit{etc}. }
\newcommand{\ie}{\textit{i.e.} }

\definecolor{MySoftBlue}{HTML}{002E90}
\definecolor{MySoftPink}{HTML}{660000}
\definecolor{LGray}{gray}{0.5}
\definecolor{grey2}{gray}{0.55}
\definecolor{grey3}{gray}{0.85}
\colorlet{mypink2}{magenta!60}

\newcommand{\OLD}[1]{#1}
\newcommand{\COMM}[1]{
{\color{grey2} #1 }
}

\newcommand{\Teach}[1]{{#1}}
\definecolor{ao(english)}{rgb}{0.0, 0.5, 0.0}
\newcommand{\Student}[1]{{\color{blue}#1}}

\definecolor{VioletBlue}{rgb}{0.2, 0.2, 0.8}

\definecolor{mygray}{gray}{.9}
\colorlet{mypink}{magenta!60}

\title{Position Aware Layer Queries for\\ Test Time Training in Vision Language Models}

\author{Rajat Modi\thanks{Correspondence to \texttt{\{rajatmodi62,priyankpathak50\}@gmail.com},\\ \texttt{yogesh@crcv.ucf.edu}}, Priyank Pathak, Xin Liang \& Yogesh Singh Rawat \\
Institute of Artificial Intelligence \\
University of Central Florida \\
Orlando, FL 32816, USA
}

\begin{document}

\maketitle
\lhead{Preprint}

\begin{abstract}
Test-Time Training (TTT) adapts models to incoming test samples (\eg out-of-distribution, (OOD)) when conventional fine-tuning is infeasible.
Existing TTT methods for Vision-Language Models (VLMs) create supervision from several augmented views, \textit{each} requiring forward (and often backward) passes through the entire VLM, incurring substantial computational cost. 
We observe that one forward pass with all the intermediate layer outputs already yields far more signal than the final embedding from all augmentations.
We introduce \textbf{Layer Query Network (LQN)}, a lightweight approach that can adapt a frozen VLM (\textit{teacher}) in a single forward pass of the VLM via a small model (\textit{student}).
LQN uses \textbf{Position-Aware Distillation (PAD)} to mimic the teacher VLM's intermediate-layer spatial tokens by \textit{querying} spatial coordinates of intermediate tokens.   
LQN additionally relies on 
\textbf{Location Consistency Regularization (LCR)}, a self-supervision technique, replacing expensive $\mathcal{O}(H{\times}W)$ image augmentation with $\mathcal{O}(1)$ coordinate sampling.  
Integrating these, LQN 
i) Adapts and improves zero-shot CLIP ViT-B/16 by 9.8\% Top-1 on OOD ImageNet, 
ii) Outperforms the previous best GS-Bias on fine-grained classification by 3.9\% Top-1, 
iii) Faster convergence than TPS for CLIP ResNet-50 (47 mins vs 55 mins)
iv) Generalizes adaptation to VLMs  like SigLIP, EVA-CLIP, and CoCa, and  lightweight students like MLP, ResNet, VGG, and 
v) Extends to panoptic, instance, and semantic segmentation.

\end{abstract}

\section{Introduction}

CLIP \citep{radford2021learning} is widely regarded as one of the first Vision-Language Models (VLMs) to demonstrate strong zero-shot generalization in tasks like image classification, image–text retrieval, 
image segmentation, video text retrieval, and audio  classification~\citep{wang2025dual, hur2025narrating, dixit2024vision}.
As real-world deployment of VLMs grows, a key question emerges:
\textit{``Is there an efficient way to boost the out-of-distribution generalization of VLMs for real-world?"}

\begin{figure*}[!t]
\centering
\includegraphics[width=0.65\textwidth]{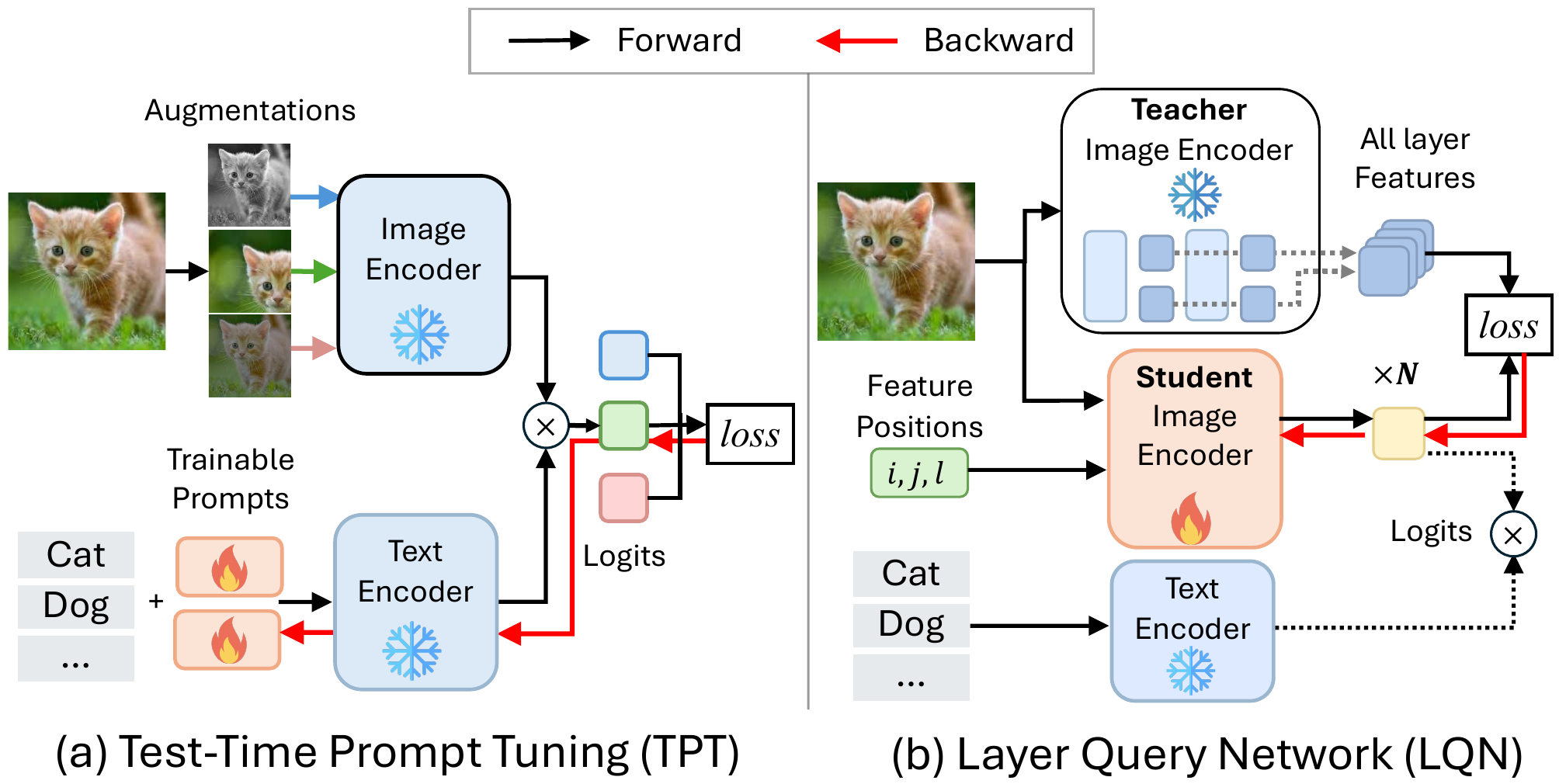}
\hfill
\includegraphics[width=0.29\textwidth]{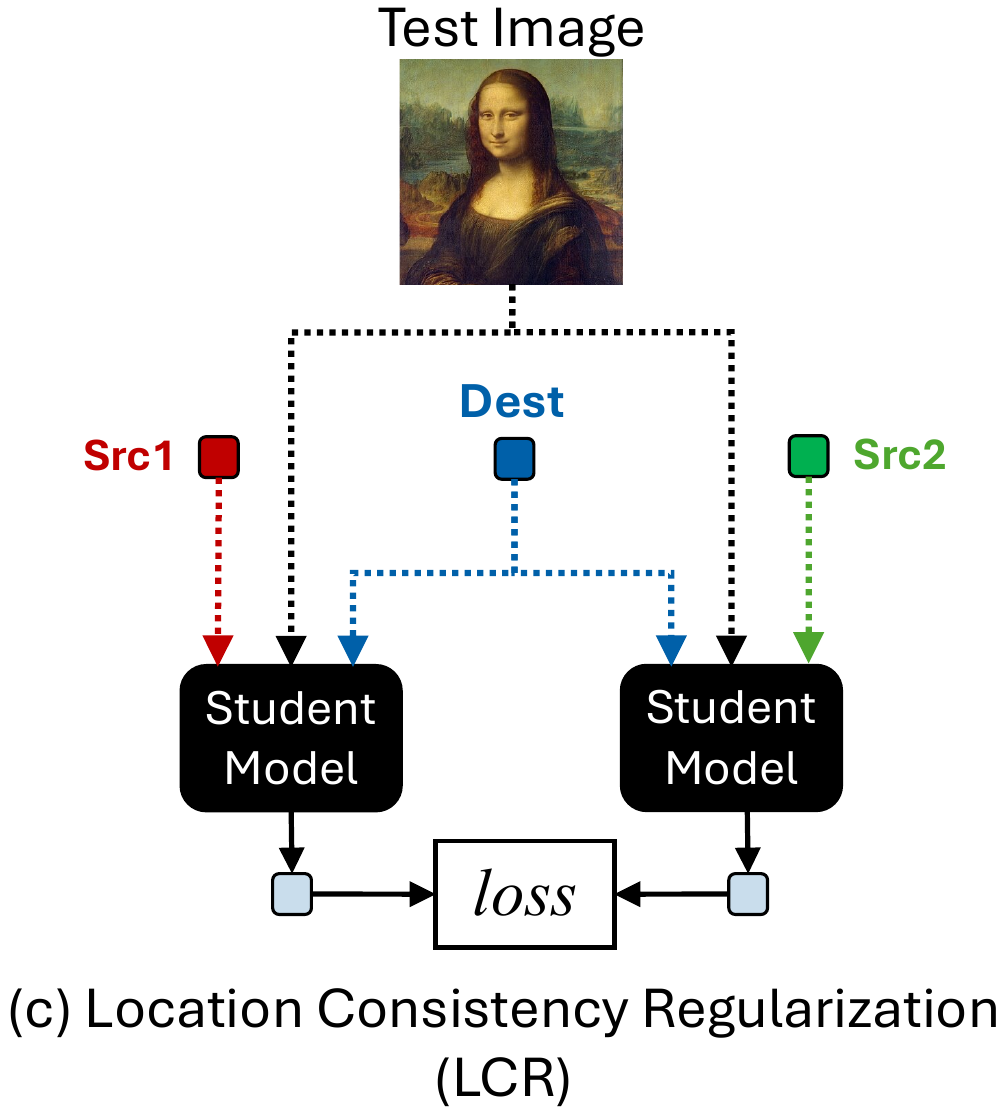}
\caption{\textbf{Comparison with existing works:} \textit{(a) Left:} TPT trains prompts via multiple augmentations, requiring  multiple forward/backward passes of VLM \textit{(b) Middle:} LQN trains small student in single forward pass from VLM (w/o any augmentation), by iteratively distilling intermediate features of image encoder via positional encoding  ($i,j,l$ coordinates for $i,j$ spatial token at $l$ layer)
\textit{(c) Right:} 
Self-supervision via varying $src$ coordinates saves compute on expensive image augmentation and avoids over-reliance on 
VLMs (teacher) embedding on OOD samples.
}
\label{fig:compare}
\end{figure*}

VLMs degrade under distribution shift, \ie unseen out-of-distribution (OOD) test domains~\citep{shu2023clipood}.
Parameter-efficient fine-tuning, such as Adapters ~\citep{yin2023adapter}, LoRA~\citep{hu2022lora}, VPT~\citep{jia2022visual}, NN \& TK0~\citep{pathak2026robust}, \etc  
can adapt VLMs but require labeled target data; an assumption that rarely holds in the real world.
Improving this, \textit{Test-Time Adaptation} `peeks' at target data on-the-fly~\citep{zhang2024dual}, but relies on multiple test samples (\eg cache seen samples) to refine predictions \citep{nguyen2025adaptive}, limiting applicability in data-constrained scenarios 
\eg medical diagnostics.
Overcoming this constraint, \underline{\textbf{T}}est-\underline{\textbf{T}}ime \underline{\textbf{T}}raining (\textbf{TTT},~\cite{sun2020test}) adapts using \textbf{just one test sample}.
TTT frameworks like TPT \citep{shu2022test} and TPS \citep{sui2025just} adapt by self-supervising 
\textit{consistency across 
multiple augmentations} for each test sample, (\cref{fig:compare} (a)). 
Although effective, this requires the VLM to perform multiple unnecessary forward (\& backward) passes for each augmentation, incurring high computational cost.

We take a different view: the signal is already \emph{inside} the VLM even without augmentations.
A single forward pass produces not only a final embedding but also $L$ layers of $H{\times}W$  spatial tokens of $D$-dimension (\eg CLIP).
We treat this stack as a \emph{3D hidden-state volume} as \textit{Teacher} output $\Teach{Y^{1,2...L}_{img}}\in\mathbb{R}^{H\times W\times L\times D}$ and fit a lightweight, randomly initialized small student network. 
Conditioned on the coordinates of the spatial tokens, this student mimics the teacher's feature at any spatial position and depth. 
Such coordinate-conditioned very small networks have the potential to 
fit smooth structure, avoiding noise~\citep{rahaman2019spectral,ulyanov2018deep}, thereby, potentially acting as a denoiser for teacher's OOD features.
We call this \textbf{Layer Query Network (LQN)}, a technique for adapting large frozen VLMs using a lightweight network in \textit{only 1 VLM forward pass} per test sample.

LQN is composed of two complementary objectives:
\textbf{Position-Aware Distillation (PAD)} and \textbf{Location Consistency Regularization (LCR)}.
PAD 
distills the teacher VLM's token at randomly sampled coordinates ($dest$, $(i,j,l)$), encoded with a 3D RoPE (\cref{fig:compare} (b)).
Unlike traditional layer-wise distillation~\citep{romero2015fitnets}, 
PAD does not need block-to-block correspondence between teacher and student, but the student can mimic every layer of a VLM.
LCR is a self-supervised regularization on the student network without teacher supervision (\cref{fig:compare} (c)). 
By varying the input spatial coordinates $src$, the student maintains consistent predictions across locations.  
Sampling the $src$ coordinate costs $\mathcal{O}(1)$, compared with expensive $\mathcal{O}(H{\times}W)$ image augmentations.
Additionally, LCR reduces the \textit{reliance} on the teacher for OOD inputs where VLM outputs may be unreliable.

LQN offers several benefits:
\textbf{1) Efficient}: Distilling intermediate teacher layers helps small student networks converge in fewer iterations (reduced GFLOPs), surpassing existing TTT approaches. 
\textbf{(2) Teacher Adaptation:} LQN outperforms existing TTT approaches in adapting CLIP for ImageNet OOD variants and fine-grained benchmarks. 
\textbf{(3) Dense prediction:} 
Intermediate feature distillation 
helps the student learn fine-grained nuances of the teacher and improve dense prediction tasks.
\textbf{(4) Self-Supervision}: Reduced reliance on the VLM, where teacher predictions may not be certain for OOD samples.   
\textbf{(5) Generalization}
Since the input rely on coordinates, LQN can predict arbitrary intermediate features, making it easy to apply to a variety of VLMs and student networks.

In summary, Layer Query Network (\textbf{LQN}) is an efficient test-time training framework for \textit{on-the-fly} adaptation of large VLMs using lightweight models without needing any image augmentations. 
LQN is evaluated on 14 classification benchmarks and 5 segmentation tasks.  
Key contributions include: 
(1) We introduce \textbf{PAD}, a position-aware distillation objective that fits all intermediate layers with one shared network, and \textbf{LCR}, a teacher-free anchor-consistency objective that replaces augmentation at $\mathcal{O}(1)$ cost.
(2) On classification benchmarks, LQN improves CLIP ViT-B/16 and ViT-L/14 on ImageNet OOD variants by 9.8\% and 5.7\% over zero-shot, and outperforms the strongest TTT baselines by 3.9\%  on fine-grained classification benchmarks.
(3) LQN extends to dense segmentation prediction by adapting EoMT on COCO, Cityscapes, and ADE20K
(4) Generalization to VLMs like EVA, SigLIP, and CoCa, and lightweight students like ResNet \& VGG.

\section{Related Work}

\textbf{VLM Adaptation:} VLMs are often adapted by fine-tuning the entire model or via LLM prompts~\citep{pratt2023does,ren2023chatgpt}. Inspired by parameter-efficient approaches~\citep{lester2021power}, adapter-based and cross-modal adaptation works have emerged~\citep{gao2023clip, lin2023multimodality}; however, gradients still flow through the \textit{entire} VLM, increasing optimization cost.

\textbf{Test-Time Optimization:} Test-Time
Adaptation (TTA) methods typically need \textit{multiple} test samples \textit{simultaneously} for stable adaptation \citep{r1,r3,r5,r7,Noori2025TestTimeAO}, while relying on historical information from multiple test samples. 
Test-Time Training (TTT) approaches like TPT \citep{shu2022test} and prompt-tuning \citep{zhou2022learning,zhou2022conditional} enforce consistency across augmented views but require costly passes through the text encoder or require annotated training data.
TPS \citep{sui2025just} avoids this by adjusting pre-computed feature vectors instead. MTA \citep{zanella2024test}, GS-Bias \citep{huang2025gs} further improve zero-shot robustness via mean-shift augmentation, and logit biasing, respectively.
TTT-MAE\citep{ttt_mae} uses SSL tasks like rotation/masked prediction, while Diff-TPT~\cite{feng2023diverse} generates OOD data via Stable Diffusion. 
In contrast to APM~\citep{modi2024apm}, 
we distill the teacher VLM's intermediate features into the student network, without relying solely on the teacher's features. 
Compared to previous works, 
LQN avoids dataset-specific pre-training, multiple expensive augmentations, and the VLM's separate forward / backward passes per augmentation, thereby saving compute.

\begin{figure*}
\centering
\includegraphics[width=0.99\textwidth]{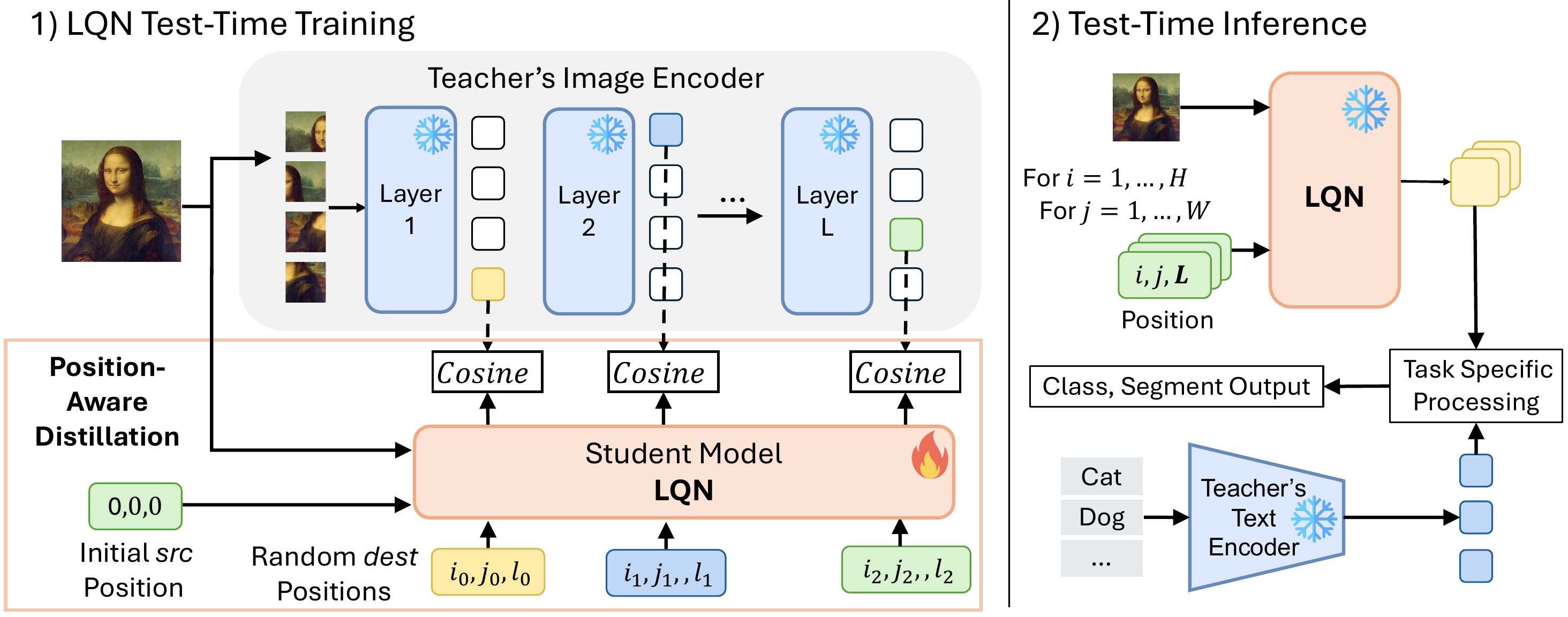}
\caption{
\textbf{Layer Query Network (LQN)} 1) For a single test image, \textit{frozen} teacher (VLM) extracts intermediate features across all layers. During \textit{test-time training}, LQN runs for $N$ iterations by randomly sampling different destination \textit{(dest)} positions $(i,j,l)$, where $(i,j)$ is the spatial location and $l$ the layer depth. Position-Aware Distillation (\textbf{PAD}) takes the image, $(0,0,0)$ as the \textit{src} position, and trains student network to predict the teacher’s feature at \textit{dest}.  
Location Consistency Regularization (\textbf{LCR}) self-supervises LQN to produce consistent features across random \textit{src}(s). 
2) After the final iteration (test-time inference), the student is queried at all spatial locations of the last layer $L$. Resultant features are processed by task-specific modules (eg, zero-shot classification/ segmentation).
}
\label{fig:main_figure}
\end{figure*}

\section{\textbf{L}ayer \textbf{Q}uery \textbf{N}etwork (LQN)}

We introduce LQN, a \textbf{T}est-\textbf{T}ime \textbf{T}raining (TTT) framework designed to enhance the out-of-distribution generalization of Vision-Language Models (VLMs).  First, we discuss the background on VLMs such as CLIP. Next, we discuss the \textbf{Position-Aware Distillation (PAD)} and \textbf{Location Consistency Regularization (LCR)}, which describes how LQN functions (\cref{sec:lqn}).

\subsection{Preliminaries}
\label{sec:Preliminaries}

A VLM like CLIP is typically pretrained on millions of image and text pairs. It consists of two parallel encoders, an image encoder $\mathcal{T}_{img}$ and a text encoder $\mathcal{T}_{text}$. 
Given an out-of-distribution test image $x_{ood}$, and a class label description $cls$, corresponding class probabilities are produced as: 
\begin{align}
Y_{\text{img}} = \mathcal{T}_{\text{img}}(x_{\text{ood}}), \quad Y_{\text{text}} = \mathcal{T}_{\text{text}}(\text{cls}), 
\quad
\mathbb{P}_{\mathrm{CLIP}}(y = c \mid (x_{\text{ood}}) = \frac{\exp(Y_{\text{text}}^c \cdot Y_{\text{img}}/t)}{\sum_{c'} \exp(Y_{{text}}^{c'} \cdot Y_{\text{img}}/t)}
\label{eq:clip_basic}
\end{align}

where, $Y_{img}$ and $Y_{text}$ are vision and text embeddings, respectively, $t$ is the temperature parameter in softmax and $\mathbb{P}_{\mathrm{CLIP}}(y = c \mid x_{ood})$ is the normalized probability  corresponding to this class label $c$. The max probability logit over $C$ classes is chosen as the class prediction.

\subsection{Test-Time Training (TTT) }
\label{sec:lqn}

Following the standard TTT paradigm~\citep{shu2022test, sui2025just}, our goal is to adapt a model (VLM) to an \textbf{unseen test sample $x_{ood}$ within $N$ iterations}. 
Conventionally, VLMs require multiple augmentations of $x_{ood}$, requiring multiple expensive forward/backward passes through the entire VLM. After adaptation, \textbf{weights are then reset before seeing the \textit{next} test sample}, preventing information leakage across test samples.
Our TTT framework adopts a teacher-student distillation framework (\cref{fig:main_figure}, \cref{alg:blind_algorithm}), where a lightweight trainable student (\eg MLP) mimics a frozen VLM teacher (\eg CLIP) for $N$ iterations.  
Repeated forward / backward passes of \textit{small} student are computationally much cheaper than those of the teacher VLM, requiring \textit{only one} forward pass of the teacher ($Y_{img}$, and $Y_{text}$, \cref{eq:clip_basic}).
Student network is optimized to mimic the teacher's intermediate spatial tokens via \textbf{Position-Aware Distillation (PAD)}, and \textbf{Location Consistency Regularization (LCR)}.
Finally, the student uses the teacher text embedding $Y_{text}$ to generate the class logit.  

\subsubsection{LQN Student Network}

Adapting / fine-tuning large-scale VLMs on \textit{just one sample} risks overfitting and potentially losing VLM's zero-shot abilities~\citep{pathak2026robust, chen2024robustsam}.
Instead, we consider \textit{adaptation by proxy},
where a small lightweight student can iteratively ($N$ iterations) adapt on a single test sample $x_{ood}$ (instead of large VLM). 
This student can  `learn' to mimic a larger VLM's internal embeddings.

Student network has three inputs: a test image $x_{ood}$, \textit{src}, and \textit{dest}  coordinates of the internal spatial features of the VLM teacher. 
For example, 3D Coordinates like (4,5,2) represent the spatial coordinate $h,w = (4,5)$, at the depth of $2^{nd}$ layer in the VLM teacher. 
\textit{`Dest'} represents the location of the target spatial token in the teacher for distillation.  
The student is tasked to maintain a consistent output despite varying \textit{input src} coordinates. 
This step may be considered equivalent to having multiple \textit{input image augmentations}. 
While augmentation is computationally costly, O($H\times W$) pixels augmentation, sampling source coordinates is O(1) retrieval, thereby helping the model generate robust embedding efficiently.
To \textit{distinguish} between the locations of $src$ and $dest$   
(line 9 of \cref{alg:blind_algorithm}), we concatenate $[\cdot \vert \cdot]$ an additional constant sinusoidal positional-encoding `flag'  $P^T\in\mathbb{R}^{2\times D}$.
$P^T[0] \text{ and } P^T[1]$ indicates location of $src  \text{ and } dest$, respectively:
\begin{equation}
\textnormal{src}= [ \textnormal{P}_{i_s,j_s,l_s} | \textnormal{P}^\textnormal{T} [0] ], \quad \textnormal{dest} = [ \textnormal{P}_{i_d,j_d,l_d} | \textnormal{P}^\textnormal{T} [1] ]
\label{eq:src_positional_encoding}
\end{equation}

\begin{algorithm}[t!]{
\small
\caption{\textbf{LQN Pseudo-Code}: Image classification shown, but can generalize for other tasks.}
\label{alg:blind_algorithm}
\textbf{Input}: OOD image $x_{ood}$, Class Labels $cls$(s) 
\begin{algorithmic}[1]

\REQUIRE Teacher Image / Text Encoder $\mathcal{T}_{img}$ / $\mathcal{T}_{text}$, 
Student LQN \Student{$\mathcal{S}_{img}$}, $N$ iterations \\
\STATE  $\Teach{Y^{1,2...L}_{img}} \gets \Teach{\mathcal{T}_{img}} (x_{ood}) \in \mathbb{R}^{H \times W \times L \times D}$ \hfill \COMM{// Teacher's all `L' layers spatial image features}
\STATE  \OLD{$\Teach{Y_{text}} \gets \Teach{\mathcal{T}_{text}} (cls)$} \hfill \COMM{// Teacher's  text features}
\STATE \OLD{$\mathcal{S}_{img} \gets N(\mu, \sigma)$} \hfill \COMM{// Random initialization of student model before seeing $x_{ood}$ test sample}
\FOR{ \OLD{iteration  $k$ in $N$ }}  
\STATE \textbf{\#\#\#\#\#\#\# Position-Aware Distillation (PAD)}
\STATE $i_d,j_d,l_d \gets$ Sample(H,W,L) \hfill \COMM{// Uniform Sample `dest' coordinates $\in [H,W,L]$}
\STATE $P_{i_d,j_d,l_d} \gets P(i_d,j_d,l_d)$ \hfill \COMM{// RoPE Positional embedding of dest}
\STATE dest $\gets [P_{i_d,j_d,l_d} | P^T [1]]$ \hfill \COMM{// Concat constant sinusoidal flag $P^T[1]$ to dest}\\
\STATE src $\gets [P_{0,0,0} | P^T [0]]$ \hfill \COMM{// For const src $(0,0,0)$
concat constant sinusoidal flag $P^T[0]$}
\STATE $L_{PAD} \gets 1 - 
\left\langle 
\frac{\Student{\mathcal{S}_{img}(src, dest, x_{ood})} \cdot \Teach{Y_{img}}[i_d,j_d,l_d]}{\|\Student{\mathcal{S}_{img}(src, dest, x_{ood})}\|_2 \|\Teach{Y_{img}}[i_d,j_d,l_d]\|_2} \right\rangle$ \hfill \COMM{// \Student{$S_{img}$} mimics teacher's `dest' features}
\vspace{5pt}
\STATE \textbf{\#\#\#\#\#\#\# Location Consistency Regularization (LCR) }
\STATE ($i_{1},j_{1},l_{1}$)  \& ($i_{2},j_{2},l_{2}$) $\gets$ Sample(H,W,\textbf{L}) \hfill  \COMM{// Sample two sources `src1' \& `src2' like lines 6-9}
\STATE src1 $\gets [P_{i_1,j_1,l_1} | P^T [0]]$ \& src2 $\gets [P_{i_2,j_2,l_2} | P^T [0]]$ \hfill \COMM{// concat flag $P^T [0]$ to src1 and src2} 
\STATE \COMM{// \Student{$S_{img}$} self-supervises to predict `Dest' features from separate src1/src2}
\STATE $L_{LCR} \gets 1 - \left\langle \frac{ \Student{\mathcal{S}_{img}(src1, dest, x_{ood}}) \cdot \Student{\mathcal{S}_{img}(src2, dest, x_{ood}}) }{ \|\Student{\mathcal{S}_{img}(src1, dest, x_{ood})}\|_2 \|\Student{\mathcal{S}_{img}(src2, dest, x_{ood})}\|_2 } \right\rangle_{\text{cos}} - \left\langle \log \|w_{src}\|_2 + \epsilon \right\rangle$
\STATE Loss $\gets L_{PAD} + \alpha L_{LCR}$ \quad $\implies$ \quad  Update $\Student{S_{img}}$\hfill \COMM{// Update using both losses} 
\ENDFOR
\STATE \textbf{ \#\#\#\#\#\#\# Prediction / Inference / Evaluation}
\FOR{$\forall (i,j) \in (H,W)$, last layer $L$} 
\STATE $P_{i,j,L} \gets P(i,j,L)$ \quad 
Dest $\gets [P_{i,j,L} | P^T [1]]$  \& Src $\gets [P_{0,0,0} | P^T [0]]$ \hfill \COMM{// Constant source $(0,0,0)$}
\STATE 
$\Student{Y^{Student}_{img}}$ += $\Student{\mathcal{S}_{img}( src, dest, x_{ood})}$ / $(H\cdot W)$ \hfill \COMM{// Spatial averaged of $x_{ood}$ features via $S_{img}$} 
\ENDFOR
\STATE  $P_{cls} \gets 
\Student{Y^{Student}_{img}}
\cdot Y_{text}$  \hfill \COMM{ // Classification is product of student image features with teacher text features } 
\STATE \textbf{Output:} $P_{cls}$
\end{algorithmic}
}
\end{algorithm}

\subsubsection{\textbf{P}osition-\textbf{A}ware \textbf{D}istillation (\textbf{PAD})}
Classically, knowledge distillation focuses on matching the final-layer features of the teacher~\citep{shu2022test, wang2021distilling}. 
Additionally, distilling large networks ($>$200M params) to small networks ($<$30M params) often requires distilling intermediate layers~\citep{hao2023one, 10678046},
thereby improving dense-task predictions (\eg segmentation).
However,  intermediate feature distillation often requires explicit correspondences between teacher and student blocks, \eg block 1 features of the student mimic block 1 of the teacher; thereby constraining the student architecture.

To relax the above constraint, we propose \textit{Position-Aware Distillation} (PAD). We collect intermediate spatial tokens across \textit{all the layers} of VLM,  $Y_{img}^{1,2,...L}$, where $Y_{img}^l \in \mathbb{R}^{H\times W \times D}$; $L$ denotes all the layers of teacher VLM, and $(H,W)$ are the all spatial tokens on layer $l$, with dimension of $D$. 
For the input test sample $x_{ood}$, the student $S_{img}$ mimics the teacher's $(i,j)$ spatial token on depth $l$, \ie $S_{img}(x_{ood}, i, j, l) \rightarrow f$, where $f\in \mathbb{R}^{D}$ is the predicted spatial vector at position $(i,j,l)$. 
For $N$ iterations, the student randomly selects (uniform sampling) a batch of these spatial indices $(i,j,l)$ and distills teacher embeddings on these indices.
However, directly feeding 3D-position coordinates  $(i,j,l)$ to a neural network leads to poor convergence~\citep{mildenhall2021nerf}. Therefore, we encode location $(i,j,l)$ as a constant 3D RoPE positional-encoding $P(i,j,l)$~\cite{su2023roformer}. 
The student takes as input a \textit{fixed} source ($src$) location $(0,0,0)$ and a \textit{random} destination $dest$ location $(i_d,j_d,l_d)$, and learns to mimic teacher's token at $dest$  via Cosine loss 
(\cref{alg:blind_algorithm} (line 10)).
\begin{align}
\textnormal{L}_\textnormal{PAD} = 1 - \left\langle \frac{ S_{\textnormal{img}}( \textnormal{src} , \textnormal{dest}, x_{\textnormal{ood}}) \cdot \textnormal{Y}_{\textnormal{img}}[i_d,j_d,l_d] }{ \| S_{\textnormal{img}}( \textnormal{src} , \textnormal{dest}, x_{\textnormal{ood}}) \|_2 \| \textnormal{Y}_{\textnormal{img}}[i_d,j_d,l_d] \|_2 } \right\rangle_{\text{cosine}}
\label{eq:bind_loss_cosine}
\end{align}

\subsubsection{\textbf{L}ocation \textbf{C}onsistency \textbf{R}egularization (\textbf{LCR})}

Traditional TTT has shown the benefits of self-supervision consistency across multiple augmented views of the image. 
Instead of expensive $\mathcal{O}(H{\times}W)$ image augmentations, we vary input coordinates of the spatial tokens, \ie sampling the $src$ coordinates  ($\mathcal{O}(1)$ cost).
We name this \textit{Location Consistency Regularization (LCR)}, a self-supervised objective that complements PAD. 
Intuitively, LCR encourages the student to produce consistent representations at a given \textit{dest} across different \textit{src(s)} locations, providing an additional signal beyond direct distillation (\cref{fig:compare} (c)).
LCR is not an augmentation, but rather a coordinate-conditioned consistency constraint that serves as a proxy for augmentation consistency.
In addition to serving as a cost-effective proxy to input augmentations, LCR
reduces the blind reliance on teacher for OOD inputs where VLM outputs may be unreliable.

Here, we sample  two random locations: $src1 = (i_1,j_1,l_1), src2 = (i_2,j_2,l_2)$ and a \textit{single} destination location $dest= (i_d,j_d,l_d)$.
The intuition is that, irrespective of whether the student operates on $(src1,dest)$ or $(src2, dest)$, it should predict the same representation at $dest$ every time
\begin{align}
\textnormal{L}_\textnormal{LCR} = 1 - \left\langle \frac{ S_{\textnormal{img}}( \textnormal{src1} , \textnormal{dest}, x_{\textnormal{ood}}) \cdot \textnormal{S}_{\textnormal{img}}( \textnormal{src2} , \textnormal{dest}, x_{\textnormal{ood}}) }{ \| S_{\textnormal{img}}( \textnormal{src1} , \textnormal{dest}, x_{\textnormal{ood}}) \|_2 \| \textnormal{S}_{\textnormal{img}}( \textnormal{src2} , \textnormal{dest}, x_{\textnormal{ood}}) \|_2 } \right\rangle_{\text{cosine}} - \left\langle \log \| w_{\textnormal{src}} \|_2 + \epsilon \right\rangle
\label{eq:recirculation_loss_reg}
\end{align}

A trivial solution to the above is the model simply ignoring the source,
\ie weight of the source  $w_{\textnormal{src}} \!\rightarrow\!0$. 
We prevent this by adding 
an extra regularization (-log with $\epsilon\!=\!1e-7$, clamped at max value of 0.7) on
$w_{\textnormal{src}}$ to prevent weight collapse. 
While PAD \textit{only} needs $dest$ ($src \leftarrow 0$), LCR uses randomly sampled $src$(s).

\subsubsection{Inference and Loss}

Both \textit{Position-Aware Distillation (PAD)} and \textit{Location Consistency Regularization (LCR)} are \textit{jointly} performed on a test sample for $N$ iterations. 
We further control the impact of LCR via $\alpha$.   
Combining both (line 16 in algorithm~\ref{alg:blind_algorithm}), the final loss is given by:
\begin{align}
\textnormal{Loss} = \textnormal{L}_\textnormal{PAD} + \alpha \textnormal{L}_\textnormal{LCR}.
\label{eq:total_loss}
\end{align}

During evaluation, features corresponding to the \textit{last-layer} $L$ of the VLM teacher is predicted; \ie $src=(0,0,0)$ and $dest = (i,j, L)$, for all $1\leq i \leq H, 1\leq j\leq W$, generating the final $Y_{img}^{student}$.  
Classification averages spatial features followed by the teacher's textual feature dot product (line 23 in \cref{alg:blind_algorithm}). 
For image segmentation, $Y_{img}^{student}$ is directly fed to the teacher's frozen mask-head (frozen mask head + adapted LQN), where upsampled features produce the segmentation mask.

\section{Experiments on LQN}
\noindent \textbf{Tasks and Datasets:}
LQN is evaluated on $14$ classification benchmarks and $5$ segmentation tasks.
Following TPT~\citep{shu2022test}, LQN is evaluated on two types of classification benchmarks: \textbf{1) Natural distribution shift}: evaluation on ImageNet val (\citeyear{imagenet}), along with its distribution-shifted variants, namely ImageNet-A (\citeyear{hendrycks2021natural})
, ImageNet-V2 (\citeyear{v2}), ImageNet-R (\citeyear{r}), and ImageNet-Sketch (\citeyear{sketch}). 
\textbf{(2) Cross-dataset generalization}: Experiments on 9 datasets, including Flowers102 (\citeyear{flowers102}), DTD (\citeyear{dtd}), OxfordPets (\citeyear{oxfordpets}), UCF101 (\citeyear{ucf101}), Caltech101 (\citeyear{caltech101}), Food101 (\citeyear{food101}), SUN397 (\citeyear{sun397}), FGVCAircraft (\citeyear{fgvcaircraft}), and EuroSAT (\citeyear{eurosat}).  
\textbf{3) Dense Segmentation task:} Results on COCO (\citeyear{coco}), ADE20K (\citeyear{ade20k}), and Cityscapes (\citeyear{cityscapes}).

\begin{table*}[t!]  
\centering  
\caption{\textbf{Natural distribution shifts:} Top-1 Results for ImageNet and its distribution-shifted variants (ImageNet-A/-V2/-R/-Sketch). \textit{Requirements} column specifies resources needed during test-time; 
\textit{Augment.}: multiple augmented views of each test sample; 
LQN doesn't need any augmented views for test samples. 
Use of the CLIP variant as a teacher is shown on the left. 
Avg: across all cols, OOD Avg: all cols except ImageNet.
}
\setlength\tabcolsep{3pt}
\resizebox{\linewidth}{!}{\begin{tabular}{c|l*{6}{c}|cc}
\specialrule{0.5pt}{0pt}{0pt}
\rowcolor{grey3} 
\multicolumn{2}{c}{Method}
& Augment. & ImageNet & ImageNet-A & ImageNet-V2 & ImageNet-R & ImageNet-Sketch & Avg & OOD Avg \\
\specialrule{1pt}{0.5pt}{0.5pt}
\multirow{10}{*}{\rotatebox[]{90}{CLIP-B/16}}  
& CLIP-ViT-B/16           & \ding{55} & 66.7 & 47.8 & 60.8 & 73.9 & 46.0 & 59.0 & 57.1 \\

& Ensemble             & \ding{55} & 68.3 & 49.8 & 61.8 & 77.6 & 48.2 & 61.1 & 59.4 \\
& TPT
&  \ding{51} & 68.9 & 54.7 & 63.4 & 77.0 & 47.9 & 62.4 & 60.8 \\

& Diff-TPT
& \ding{51} & 70.3 & 55.6 & 65.1 & 75.0 & 46.8 & 62.6 & 60.6 \\

& MTA + TPT
& \ding{51} & 70.0 & 58.0 & 64.2 & 78.3 & 49.6 & 64.0 & 62.5 \\
& APM
& \ding{55} & 68.1 & 52.1 & 67.2 & 76.5 & 49.3 & 62.6 & 61.3 \\

& GS-Bias
& \ding{51} & 70.5 &56.6 & 64.6 & 80.4 &50.3  & 64.5  & 63.0 \\
& TPS
& \ding{51} & 70.1 & 60.0 & 64.7 & 80.2 & 49.9 & 65.0 & 63.7 \\

\hhline{~---------}
\rowcolor{black!5}
\rowcolor{black!5}
& LQN \textit{(Our)}
& \ding{55} & 73.1 & 60.8 & 71.2 & 83.1 & 52.6 & 68.2 & 66.9 \\
\rowcolor{black!5}
& LQN (Pretrained, \textit{Our}) 
& \ding{55} & 74.5 & 62.3 & 72.9 & 84.8 & 53.9 & 69.7 & 68.5 \\

\hline

& CLIP ViT-L/14          & \ding{55} & 76.2 & 69.6 & 72.1 & 85.9 & 58.8 & 72.5 & 71.6 \\
& APM
& \ding{55} & 77.3 & 71.8 & 72.8 & 87.1 & 62.2 & 74.2 & 73.5 \\
\hhline{~---------} 

\rowcolor{black!5}
& LQN \textit{(Our)}
& \ding{55} & 80.5 & 75.4 & 76.9 & 91.3 & 65.7 & 78.0  & 77.3 \\
\rowcolor{black!5}
\multirow{-4}{*}{\rotatebox[]{90}{CLIP-L/14}} & 
LQN (Pretrained, \textit{Our}) 
& \ding{55} & 81.4  & 76.5 & 77.5 & 92.2  & 66.4 & 78.8 & 78.2 \\
\specialrule{1pt}{0.5pt}{0.5pt}
\end{tabular}
}
\label{tab:imagenet_splits}
\end{table*}

\noindent \textbf{Evaluation Metrics}
For experiments with distribution shifts (Tab\ref{tab:imagenet_splits}, \ref{tab:lqn_ood_vlms}), we report Top-1 classification accuracy (across all datasets, namely ImageNet val, ImageNet-A, ImageNet-V2, ImageNet-R, ImageNet-Sketch), and Top-1 OOD Accuracy (all splits except ImageNet val). For dense segmentation tasks (Tab \ref{tab:combined_segmentation}), we report Panoptic Quality (PQ), and Average Precision.

\noindent \textbf{Implementation}
LQN student $(25M)$ contains $5$ linear layers, with ReLU activation and one conv filter of $16 \times 16$ kernel, and stride $16$.   
We evaluate two variants of  LQN:  LQN and LQN (Pretrained).
(i) \textbf{LQN} uses PAD and LCR and is initialized from scratch. (ii) \textbf{LQN (Pretrained)} first performs SSL pre-training (distilling) ImageNet features from DINOv2, and then perform TTT on each sample independently. 
We \textit{reset} the weights to pre-trained weights prior to processing a new sample.   
Detailed implementation details in Supplementary~\cref{sec:implementation}.

\subsection{Main Results}

\begin{table*}[t]
\caption{\textbf{Cross-dataset generalization from ImageNet to fine-grained classification tasks.} 
Top-1 accuracy reported. 
CLIP-B/16 as teacher.
Avg. averages all columns. 
}
\centering
\setlength\tabcolsep{3pt}
\resizebox{\linewidth}{!}{\begin{tabular}{l*{10}c | c}
\specialrule{1pt}{0pt}{0pt}
\rowcolor{grey3} 
Method      & Augment.    & Flower102   & DTD  & Pets  & UCF101   & Caltech101  & Food101 & SUN397 & Aircraft & EuroSAT & Avg \\     
\specialrule{1pt}{0.5pt}{0.5pt}

CLIP-ViT-B/16   & \ding{55} & 67.4   & 44.3  &  88.3    &  65.1   &   93.4   &  83.7   &   62.6    &   23.7    &   42.0     &    63.4     \\

Ensemble        & \ding{55} & 67.0   & 45.0  &  86.9    &  65.2   &   93.6   &  82.9   &   65.6    &   23.2    &   50.4     &    64.4      \\

TPT 
&  \ding{51} &  69.0   & 47.8  &  87.8    &  68.0   &   94.2   &  84.7   &   65.5    &   24.8    &   42.4     &    64.9      \\

DiffTPT
&  \ding{51} & 70.1   & 47.0  &  88.2    & 62.6   &   92.4   &  87.2   &   65.7    &   25.6    &   43.1     &    64.7      \\

MTA
& \ding{51} &  68.0  &  45.9 &  88.2    & 68.6   & 94.2     &  85.0  &   66.6   &   25.2    &     45.3   &     65.2     \\
APM
& \ding{55} &  62.0   & 48.9  &  81.6    &  72.6   &   89.6   &  84.2   & 65.7 &   29.7    &   55.7     &    65.5      \\
GS-Bias
&  \ding{51} &  71.9  & 46.1  &   90.3   & 67.5   &  94.6    &  86.0 &   67.4   &  26.4     &   52.4    &    67.0    \\
\specialrule{1pt}{0.5pt}{0.5pt}

\rowcolor{black!5} LQN
\textit{(Our)}    
& \ding{55} &  69.8  & 54.5  &   88.1  &  76.3   &  96.2    &  89.2   &  70.9    &    33.1   &   60.2   &       70.9   \\

\rowcolor{black!5} LQN (Pretrained, \textit{Our})
& \ding{55} &  71.1  &  55.9 &  89.4   &   77.5  &  97.1    &   90.8  & 72.4     &    34.2   &   61.8   &    72.2      \\

\specialrule{1pt}{0.5pt}{0.5pt}
\end{tabular}
}
\label{tab:fine-grained}
\end{table*}
\begin{table*}[t!]
\centering
\small
\caption{\textbf{Dense-task Segmentation} 
Input resolution of $1280\times1280$; semantic segmentation has $1024\times1024$. PQ: Panoptic Quality, AP: Average Precision. LQN performs TTT on EoMT (teacher).
}
\setlength\tabcolsep{12pt}
\resizebox{0.9\linewidth}{!}{\begin{tabular}{l 
| c c |c | c c }
\specialrule{0.5pt}{0pt}{0pt}
\rowcolor{grey3} 
& \multicolumn{2}{c|}{\textbf{Panoptic}} 
& \multicolumn{1}{c|}{\textbf{Instance}} 
& \multicolumn{2}{c}{\textbf{Semantic}} \\
\rowcolor{grey3} 
& COCO & ADE20K & COCO 
& Cityscapes & ADE20K \\
\rowcolor{grey3} 
\multirow{-3}{*}{Method}  
& PQ ↑ & PQ ↑ & AP ↑ 
& mIoU ↑ 
& 
mIoU ↑ \\
\specialrule{1pt}{0.5pt}{0.5pt}
EoMT
& 58.3 & 51.7 & 48.8 & 84.2 & 58.4 \\

\hline

\rowcolor{black!5} LQN \textit{(Our)}
& 64.5 & 57.1 & 55.2 & 87.5 & 64.3 \\

\rowcolor{black!5} LQN (Pretrained, \textit{Our})
& 64.9 & 58.6 & 56.7 & 88.2 & 65.8 \\

\specialrule{1pt}{0.5pt}{0.5pt}
\end{tabular}
}
\label{tab:combined_segmentation}
\end{table*}

\begin{table*}[!t]
\centering
\caption{\textbf{Generalization across various VLMs teacher} LQN+\{SigLIP, EVA-CLIP for various classification benchmarks. 
Similar averaging as 
\cref{tab:imagenet_splits} and 
\cref{tab:fine-grained}.
}
\label{tab:lqn_ood_vlms}
\setlength\tabcolsep{2pt}
\resizebox{\textwidth}{!}{\begin{tabular}{lccccc|cc||ccccccccc|c}
\specialrule{0.5pt}{0pt}{0pt}
\rowcolor{grey3} 
& \multicolumn{7}{c||}{ImageNet and OOD} & \multicolumn{10}{c}{Fine-grained classification} \\ 
[-.4pt]\hhline{~-----------------} 
\rowcolor{grey3} 
\multirow{-2}{*}{Method}
& ImageNet & IN-A & IN-V2 & IN-R & IN-Sketch & Avg & OOD Avg & Flowers102 & DTD & Pets & UCF101 & Caltech101 & Food101 & SUN397 & Aircraft & EuroSAT & Avg \\
\specialrule{1pt}{0.5pt}{0.5pt}
SigLIP & 76.0 & 45.3 & 68.9 & 90.3 & 67.9 & 69.7 & 68.1 & 85.8 & 64.7 & 94.1 & 72.5 & 90.5 & 89.8 & 69.8 & 43.8 & 43.8 & 72.8 \\
\rowcolor{black!5}    + LQN  & 79.2 & 48.7 & 72.4 & 93.4 & 71.6 & 73.1 & 71.5&
89.2 & 68.5 & 97.4 & 76.2 & 94.1& 93.1& 73.5 & 47.3 & 47.6 & 76.3
\\
\midrule
EVA-CLIP & 76.1 & 64.6 & 68.9 & 89.1 & 63.3 & 72.4 & 71.5 &
72.0 & 59.3 & 93.7 & 74.1 & 90.4 & 89.7 &  71.9 & 28.5 &  69.9 & 72.2 \\
\rowcolor{black!5}    + LQN    & 79.8 & 68.9 & 73.1 & 92.5 & 67.4 & 76.3 & 75.5 &
76.4 & 63.2  & 97.2 & 78.6 & 94.3 & 93.7 & 75.9 & 32.7 & 74.1 & 76.2
\\
\midrule
CoCa   & 63.6 & 21.5 & 55.7 & 73.2 & 51.3 & 53.1 & 50.4 & 
64.7 & 53.3 & 89.1 & 61.4 & 89.1 & 77.3 & 66.1 & 18.8 & 45.3 & 62.8
\\
\rowcolor{black!5}  + LQN  & 68.1 & 25.1 & 60.2 & 77.6 & 55.7 & 57.3 & 54.7 & 
69.4 & 57.8 & 93.6 & 66.0 & 93.3 & 82.1 & 70.6 & 22.9 & 50.1 & 67.3
\\
\bottomrule
\end{tabular}
}
\end{table*}

\noindent \textbf{1) Natural Distribution Shifts}: 
\Cref{tab:imagenet_splits} evaluates LQN on ImageNet and its 4 OOD variants. 
Zero-Shot CLIP underperforms, obtaining $57.1\%$ average OOD accuracy. 
Compared with other TTT baselines (\eg TPT~\citeyear{shu2022test}, Diff-TPT~\citeyear{feng2023diverse}, 
MTA + TPT, APM~\citeyear{modi2024apm}, GS-Bias~\citeyear{huang2025gs}, TPS~\citeyear{sui2025just}), LQN/LQN (pretrained) on average outperform by 4.9\%/6.5\% on OOD-Avg CLIP-ViT-Base. 
\textbf{2) Cross-Dataset Generalization:} \Cref{tab:fine-grained} observes a similar performance improvement on fine-grained benchmarks. 
LQN/LQN (pretrained) on average outperforms previous results by 5.4\%/6.8\%, and even surpasses previous best GS-Bias. 
\textbf{3) Generalization to segmentation:} Table~\ref{tab:combined_segmentation} shows the results for more challenging dense-level segmentation tasks, where LQN adapts EoMT~\citep{kerssies2025eomt} as teacher.  
\Cref{fig:qualitative_segmentation} (left) shows a few sample qualitative segmentation results of LQN. \textbf{4) Generalization across VLMs:} \Cref{tab:lqn_ood_vlms} shows LQN's generalization across various VLMs other than CLIP like SigLIP~\citep{zhai2023sigmoid}, EVA~\citep{fang2023eva}, and CoCa~\citep{yu2022coca}
5) Extension to Test Time Augmentation (TTA):
We have mainly focused on TTT; however, LQN can be generalized to TTA (Supplementary tab. \ref{tab:tda}), \ie using multiple data points. 

\subsection{Analysis on LQN}

\begin{wrapfigure}{r}{5.2cm} 
\vspace{-15pt}
\centering
\includegraphics[width=\linewidth]{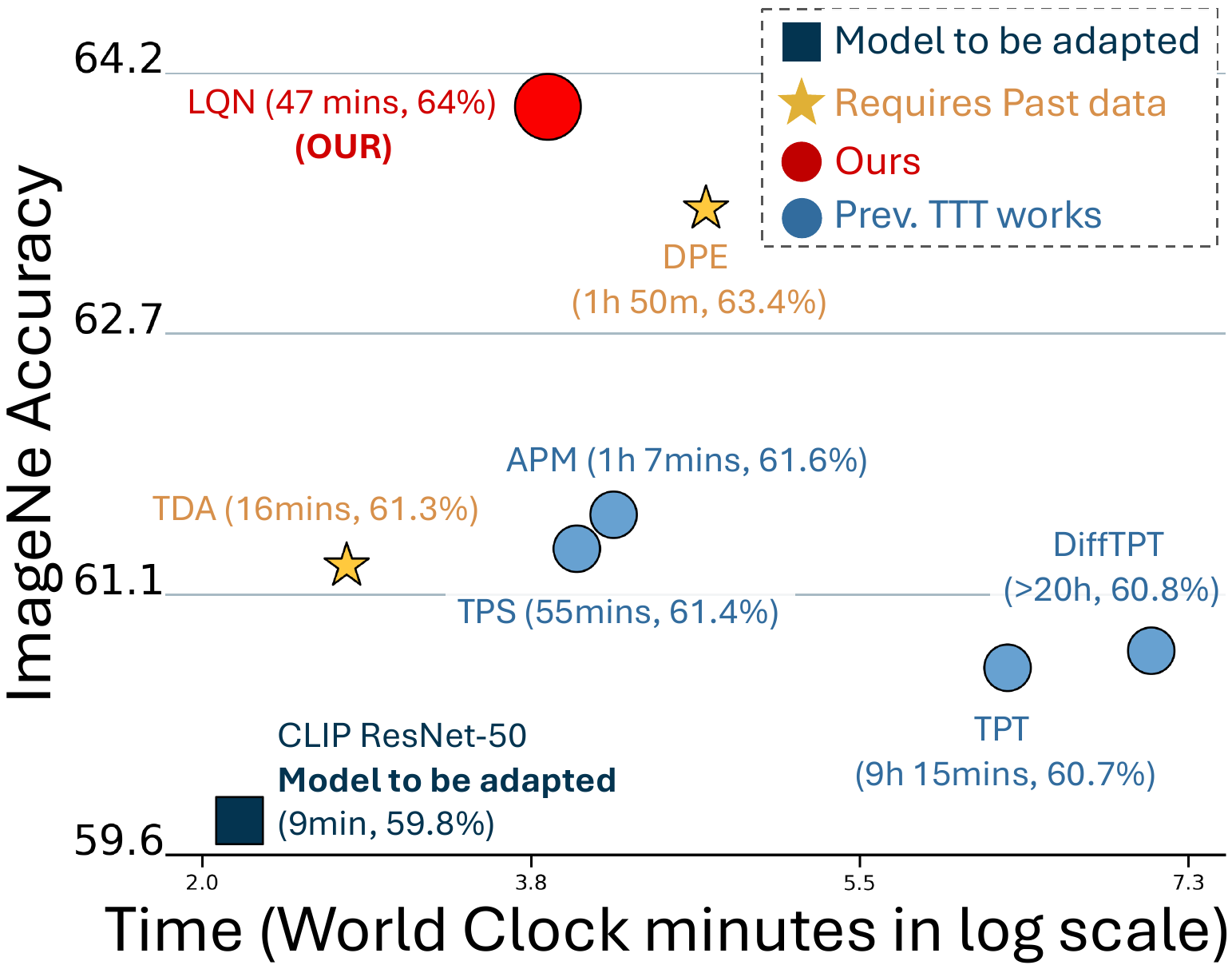}
\vspace{-10pt}
\caption{\textbf{Wall-clock time} CLIP ResNet-50 on ImageNet. 
}
\label{fig:clock}
\end{wrapfigure}
\begin{figure}[t]
\centering
\centering
\includegraphics[width=0.49\linewidth]{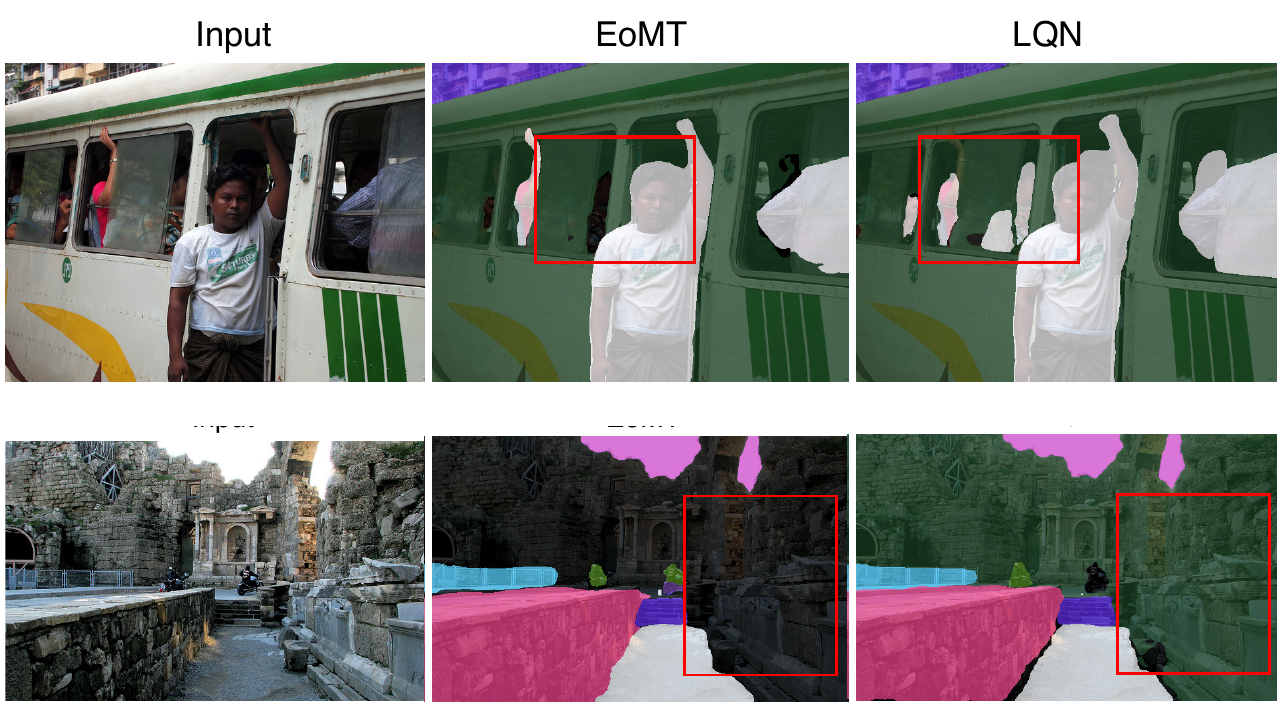}
\includegraphics[width=0.49\linewidth]{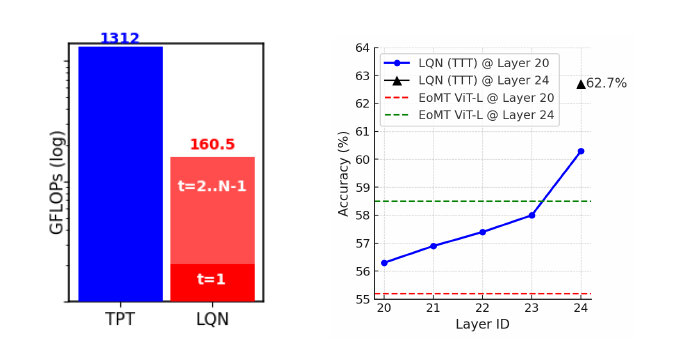}
\caption{
\textit{Left:}
\textbf{Segmentation Mask Quality: EoMT vs LQN}
(Above) LQN can even segment persons \textit{occluded} behind the window of the bus. (Bottom) LQN semantically groups visual elements of the scene (\eg walls), whereas EoMT falls short.
\textit{Middle:}
\textbf{Cost vs. Performance.} \textit{Left)} TPT’s 63 augmentations result in high overhead. \textit{Right:} \textbf{Generalization:} Performance improves even at unseen final layers (COCO PQ), showing LQN's ability to generalize beyond TTT teacher layers. 
}
\label{fig:qualitative_segmentation}
\label{fig:imagenet_efficiency_cost}
\end{figure}

\textbf{FLOP Analysis:} 
\Cref{fig:imagenet_efficiency_cost} (middle) compares total GFLOPS consumed during TTT for both TPT and LQN. 
TPT only does one gradient update per test sample; however requires $63$ augmentations, \ie $63+1=64$ forward passes through VLM, totaling $1312$ GFLOPS. 
In contrast, LQN performs one VLM forward pass (CLIP) costing $20.5$ GFLOPs. 
However, the subsequent $N=14$ backward steps for updating the student network cost $10$ GFLOPs, yielding a total of $20.5 + 14 \times 10 = 160.5$ GFLOPs; far less than TPT. 
\Cref{fig:clock}
reports the \textit{actual wall-clock time}~\citep{zhang2024dual} required for TTT over $50{,}000$ ImageNet validation images on a single A6000 GPU. APM consumes $1$hr+, while LQN converges in $47$mins; improving accuracy by $4.2\%$.

\begin{figure*}[t!]
\centering
\includegraphics[width=\linewidth]{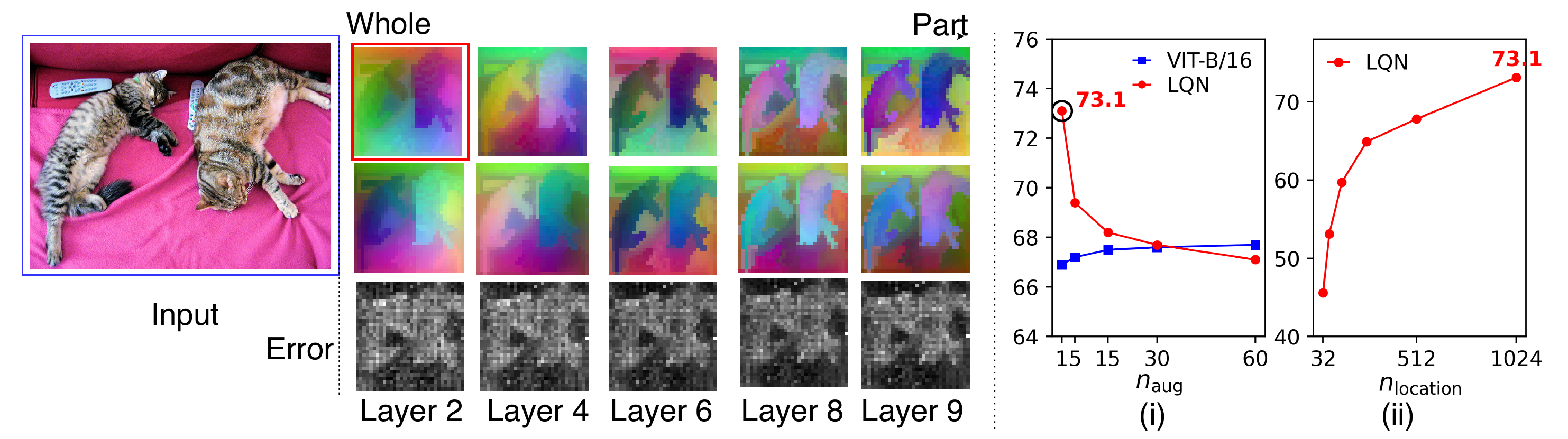}  
\vspace{-2pt}
\caption{\textit{Left:} Token visualizations of the teacher model EoMT (top), LQN (middle), and their difference (\textit{Error}, bottom) showing 
LQN’s predictions matching EoMT.
\textit{Right (i)} Augmenting test image during TTT \textit{harms} LQN (red), but improves CLIP ViT-B/16 (blue). 
\textit{(ii)} Distilling \textit{more} teacher locations improves performance, verifying benefits of intermediate layer distillation.
$n_{aug}$ \& $n_{location}$ represent the \# of augmentations and the \# of coordinates to distill in PAD. 
}
\label{fig:ablation_tokenVisualization}
\end{figure*}

\noindent \textbf{Unseen teacher layers:} In ~\Cref{fig:imagenet_efficiency_cost} (right), we use EoMT ViT-L as LQN's teacher for panoptic segmentation on COCO. When TTT is performed by distilling \textit{all} 24 layers of the teacher into LQN, we obtain a peak mIoU of $64.5$. 
Next, we perform TTT using only the \textit{first 20 layers} of EoMT. However, during inference, we predict that the $20-24$ layer as input to the segmentation head, observing a performance increase from $56.2$ in the $20^{th}$ layer to $60.3$ in the $24^{th}$ layer.
This shows that 
the position-conditioned student can extrapolate to layers not seen during PAD.

\noindent \textbf{LQN similarity with teacher's features:} We feed-forward a sample COCO dataset image into both the EoMT teacher and LQN. Instead of predicting the final-layer features, we predict intermediate-layer spatial tokens. 
Fig~\ref{fig:ablation_tokenVisualization} (left) shows the 
t-SNE of the intermediate representations of EoMT teacher and LQN prediction. 
We observe that the predicted features are very similar to the teacher's. 

\noindent \textbf{Why not augment test sample itself?} In Fig.~\ref{fig:ablation_tokenVisualization}(right)(i), we perform augmentation of a same test-sample. This improves ViT-B based model consistent with prior observations in TPT/TPS. However, this harms LQN performance. 
Fig.~\ref{fig:ablation_tokenVisualization}(right)(ii), we gradually increase the number of locations (with uniform sampling) to distill over (PAD)\footnote{
Uniform sampling from a subset of $H\times W\times L$ teacher VLM internal locations (Algo \ref{alg:blind_algorithm} line 6).}. 
We observe performance increases with more spatial coordinates, indicating the importance of distilling all locations.

\noindent \textbf{Both src \& dest vs just dest?}
\Cref{tab:lqn_variants}  
compares LQN variants with only $dest$ (no src) \ie only PAD (LCR only makes sense with src).  This has the worst performance, possibly because of \textit{blind reliance} on the teacher for $x_{ood}$.
Next, we apply LCR (\cref{eq:recirculation_loss_reg}) by enforcing self-consistency between locations $dest$ and $dest + \triangle$, where  $\triangle$ is Gaussian noise. 
The intuition is that minor perturbations in dest should self-supervise and reduce reliance on the teacher (improved performance over only PAD). 
Finally, we test our LQN with 
consistency between $(src1,dest)$ \&  $(src2,dest)$, outperforming others.
This indicates the importance of reduced reliance on the teacher, while maintaining self-consistency via \textit{src}. 
We believe this is likely because $src$ comes from a similar domain as other $dest$ input, while $\triangle$ Gaussian noise likely suffers from a domain mismatch with $dest$.

\subsection{Ablations on LQN }
\label{sec:Ablations}

We discuss various design choices made in LQN. We perform the following ablations on Food-101 classification (Top-1 Acc) and COCO segmentation (Panoptic Quality (PQ)).

\noindent \textbf{LQN Student can work with multiple architectures}: 
LQN can be generalized to standard small ConvNet networks like VGG/ResNet18/34 ($<30$ M params) as well. 
\Cref{tab:lqn_diff_architectures} shows the results for these backbones (ImageNet pretrained) as student networks. 
ResNets / VGG add an additional channel to the RGB input (3+1 channels) with the extra channel containing $d$-dimensional positional encoding repeated for all spatial locations. 
Using a trainable projector, 4D channel input is projected back to 3D input.
All other layers are kept frozen.

\noindent \textbf{Different Losses}: 
\Cref{tab:loss_comp} shows the effects of different losses applied to LQN for \textit{both} PAD / LCR. Intuitively, $L_{1}/L_{2}$ constrain the student to exactly mimic the teacher's features ($>200$ M params), which may be harder given smaller student ($<30$ M params).  
We observe that cosine performs the best, since it enforces angular similarity between student and teacher (alignment). 

\noindent \textbf{Positional embeddings}:  \Cref{tab:pos_comp} shows different strategies for injecting positional information in LQN. Simply injecting 3D integer coordinates harms performance. 
RoPE outperforms absolute encoding \citep{vaswani2017attention}, indicating that rotating vectors in higher dimensions works well.

\noindent 
\textbf{Sampling Strategy of teacher's locations}: \Cref{tab:sampling_strategies} shows results with different `spatial token location' sampling strategies, \eg uniform, higher probabilities of sampling from shallow / deeper layers. Uniform sampling appears to yield optimal results.

\noindent \textbf{LCR Regularization}:
LQN's input consists of both $src$ and $dest$.
A trivial solution for LCR (Eq. \ref{eq:recirculation_loss_reg}) is when $src$ weights $\rightarrow 0$, \ie the output 
depends only on $dest$, ignoring $src$. 
\Cref{tab:weight_penalty_ablation} shows 
weight regularization 
$\left\langle \log \| w_{\textnormal{src}} \|_2 + \epsilon \right\rangle$  
has a visible performance impact in preventing this collapse.

\begin{table}[t!]
\begin{minipage}[c]{0.28\textwidth}
\centering
\setlength\tabcolsep{2pt}
\caption{\textbf{LCR variations} 
W/o $src$, only PAD can be applied. 
$\triangle$: LCR w/ Gaussian noise instead of $src$.
}
\label{tab:lqn_variants}
\setlength\tabcolsep{2.2pt}
\resizebox{\linewidth}{!}{\begin{tabular}{lcc}
\specialrule{1pt}{0pt}{0pt}
\rowcolor{grey3} 
\textbf{Parts} & Food-101 & COCO \\ 
\specialrule{1pt}{0.5pt}{0.5pt}
PAD (-$src$)  & 71.8      & 54.9 \\
PAD + $\triangle$   &   76.0  & 58.1\\
\rowcolor{black!5} PAD + LCR & 89.2 & 64.5 \\ \specialrule{1pt}{0.5pt}{0.5pt}
\end{tabular}
}
\end{minipage}
\hfill
\begin{minipage}[c]{0.21\textwidth}
\centering
\setlength\tabcolsep{1.2pt}
\caption{\textbf{Student Architectures}.}
\label{tab:lqn_diff_architectures}
\resizebox{\linewidth}{!}{\begin{tabular}{lr}
\specialrule{1pt}{0pt}{0pt}
\rowcolor{grey3} 
Student & Food-101 \\
\specialrule{1pt}{0.5pt}{0.5pt}
VGG          & 84.9 \\
ResNet 18    & 85.4 \\
ResNet 34    & 87.2 \\
MLP         & 90.8 \\
\specialrule{1pt}{0.5pt}{0.5pt}
\end{tabular}
}
\end{minipage}
\hfill
\begin{minipage}[c]{0.21\textwidth}
\centering
\renewcommand{\arraystretch}{1.2}
\setlength\tabcolsep{1pt}
\caption{Type of loss in PAD \& LCR }
\label{tab:loss_comp}
\resizebox{\linewidth}{!}{\begin{tabular}{lcc}
\specialrule{1pt}{0pt}{0pt}
\rowcolor{grey3} 
Loss & Food-101 & COCO \\ \specialrule{1pt}{0.5pt}{0.5pt}
L1      & 75.8     & 52.0 \\
MSE     & 85.7     & 61.1 \\
\rowcolor{black!5} Cosine & 89.2 & 64.5 \\ 
\specialrule{1pt}{0.5pt}{0.5pt}
\end{tabular}
}
\end{minipage}
\hfill
\begin{minipage}[c]{0.23\textwidth}
\centering
\setlength\tabcolsep{1.2pt}
\caption{\textbf{Positional Encodings}. Coord. feeds int positions. 
}
\label{tab:pos_comp}
\resizebox{\linewidth}{!}{\begin{tabular}{lcc}
\specialrule{1pt}{0pt}{0pt}
\rowcolor{grey3} 
Pos. & Food-101 & COCO \\ 
\specialrule{1pt}{0.5pt}{0.5pt}
Coord.     & 33.1   & 27.9 \\
Sin-Cos     & 88.6     & 63.8 \\
\rowcolor{black!5} RoPE & 89.2 & 64.5 \\ 
\specialrule{1pt}{0.5pt}{0.5pt}
\end{tabular}
}
\end{minipage}
\end{table}

\begin{table}[t]
\begin{minipage}[c]{0.48\textwidth}
\centering
\setlength\tabcolsep{2pt}
\caption
{Sampling Strategies for VLM Layers.}
\resizebox{\linewidth}{!}{\label{tab:sampling_strategies}
\begin{tabular}{lcc}
\specialrule{1pt}{0pt}{0pt}
\rowcolor{grey3} 
\textbf{Sampling Strategy} & Food-101 & COCO \\
\specialrule{1pt}{0.5pt}{0.5pt}
Sampling more from shallow layers & 81.9 & 43.7 \\
Sampling more from deeper layers & 87.8 & 52.0 \\
Uniform Sampling           & 89.2 & 64.5 \\
\specialrule{1pt}{0.5pt}{0.5pt}
\end{tabular}
}
\end{minipage}
\hfill
\begin{minipage}[c]{0.46\textwidth}
\centering
\setlength\tabcolsep{2pt}
\caption{Effect of Weight Regularization (Reg.) on LCR (\Cref{eq:recirculation_loss_reg})}
\label{tab:weight_penalty_ablation}
\resizebox{\linewidth}{!}{\begin{tabular}{lcc}
\specialrule{1pt}{0pt}{0pt}
\rowcolor{grey3} 
\textbf{Configuration} & Food-101 & COCO \\
\specialrule{1pt}{0.5pt}{0.5pt}
PAD + LCR (w/o weight Reg.) & 87.4 & 62.6 \\
PAD + LCR (w/ weight Reg.)  & 89.2 & 64.5 \\
\specialrule{1pt}{0.5pt}{0.5pt}
\end{tabular}
}
\end{minipage}
\end{table}

\section{Conclusion \& Future Work}

Our work shows it is possible to adapt VLMs on unseen test samples by updating a small proxy network, without relying on expensive augmentations or multiple forward/backward VLM passes. 
\textbf{L}ayer \textbf{Q}uery \textbf{N}etwork (LQN) allows us to mimic VLMs' intermediate features conditioned on 
3D positional coordinates of the features (\textbf{P}ositional \textbf{A}ware \textbf{D}istillation, PAD), while reducing reliance on the teacher via \textbf{L}ocation \textbf{C}onsistency \textbf{R}egularization (LCR). 
Conditioning input on 3D coordinates helps generalize LQN across VLMs like CLIP, SigLIP, EVA, and CoCa, and smaller networks like MLP, ResNet, and VGG. 
Empirical generalization is shown across a variety of classification and segmentation tasks, while maintaining faster convergence compared to its counterparts.

A potential next step could be to generalize these capabilities for streaming videos, eg, adapting selectively/dynamically to incoming frames. In principle, memory/compute may be improved further by reducing the overhead of intermediate feature storage (during backprop), efficient feed-forward via top-k token selection, and feature compression.

\YSR{check for duplicate references: Yuan et al. 2023a = 2023b}

\YSR{missing from refenrences?
DiffTPT (Feng et al., ICCV 2023), ZERO (Farina et al., NeurIPS 2024), TDA (Karmanov et al., CVPR 2024), DMN-ZS, DynaPrompt, TTT-Online and UDA-SS (Sun et al.)
OpenCLIP, DINOv2 (Oquab et al.), RoPE (Su et al.), Vision Mamba (Zhu et al., 2024), EoMT's ViT backbone
ImageNet-C and CIFAR-10-C (Hendrycks  Dietterich, 2019), t-SNE, fvcore, Adam, MedCLIP, the Kaggle pneumothorax dataset
TTT-MAE is named in Related Work without a citation.
}

\YSR{are these works relevant?
Single-image VLM TTT: C-TPT (ICLR'24), RLCF (ICLR'24), WATT (NeurIPS'24), CLIPArTT, TTL, R-TPT (CVPR'25), BoostAdapter (NeurIPS'24).}

\YSR{make sure we have TENT etc in related work, reviewers only know TENT when you talk about adaptation... }

\YSR{see if we can cite these, Coordinate and implicit feature networks: FeatUp, LIIF, Distilled Feature Fields, SIREN.}
\PP{resolve these .... }

\section*{Reproducibility Statement} 
\label{sec:reproducibility}
To facilitate reproducibility, we have included the detailed implementation details and hyperparameters in the Supplementary. Supplementary \cref{sec:implementation} describes the number of neurons in each layer of LQN. Similarly, Table \ref{tab:hyperparameter} discusses the size of the hidden dimension, floating point precision, and hardware configuration of the computing machine on which experiments were performed.

\bibliography{iclr2027_conference}
\bibliographystyle{iclr2027_conference}

\clearpage
\appendix
\noindent {\huge \textbf{Appendix}}

\section{Broader Impact}
\label{sec:broader_impact}
There are two core ideas that motivated us to design the Layer Query Network (LQN).
The first idea is that positional encodings could function as an addressing mechanism \citep{vaswani2017attention}. When a network is conditioned on a specific positional encoding, it becomes capable of `retrieving' the relevant entity stored at that location \citep{modi2024apm}.
The second idea is that by "jumping" between higher and lower layer indices during LCR, we simulate both bottom-up and top-down dynamics\citep{hinton2023represent}.  

LQN can be thought to lie in the middle of two extremes of distillation literature: 1) standard knowledge-distillation only distills logits of the final layer of the teacher. This misses rich intermediate representations. 2) layer-wise knowledge-distillation distills few/all layers, however requires the student to have identical structure to teacher, leading to slow inference. LQN offers the best of both worlds: it allows weight-sharing the net across all intermediate layers, and compute features in a single forward-pass.

\textbf{Limitations:} 
While this work focused on processing one test sample at a time (TTT), in the future we would like to study how such co-ordinate based network perform when trained on large-scale data. 
Ultimately, we would like to study LQN on videos: \textit{mimicking} how spatio-temporal processing occurs in the infero-temporal pathway of human primates.

\textbf{Ethical Considerations:}  We emphasize that LQN should not be deployed in applications involving surveillance, manipulation, or deceptive content generation, and advocate for its use in settings that prioritize transparency, accountability, and societal benefit. A scenario may emerge where LQN student-teacher can learn `recursively' from each other and lead to  a form of self-referential learning. A theoretical ceiling on improvements which can be obtained/safety-alignment should be studied.  Please note that experiments with teacher also being tuned in parallel to the student remain out of scope of this paper, and instead follow standard experimental setups in TTT literature. Similarly, training such nets at large-scale should adhere to principles put forth in \citep{bengio2024managing}.

\section{Implementation Details}
\label{sec:implementation}
\textbf{Architecture:} In Tab~\ref{tab:architecture}, we inflate the full architecture of our LQN. LQN consists of a single CNN filter. Given an input image $x_{ood}$, we first run a single convolution filter on it with a stride $s$. The resultant $h/s \times w/s \times 1$ vector is then `flattened'. The two locations $(src,dest)$ are encoded as 3-D positional encodings, along with the flag positional encoding (Alg \ref{alg:blind_algorithm}). The flattened input image, along with $d$-dimensional source, and $d$- dimensional destination are passed through an MLP. The first layer of the MLP contains $(h/s*w/s + d_p + d^{\dagger}_p)$ * $4096$ learnable parameters. The subsequent layers contain $4096$ * $4096$, $4096$ * $2048$ , $2048$ * $1024$ parameters respectively. Finally, we have a projection head which projects the MLP output to $1024, d$, where $d$ is the dimensionality of internal tokens of a teacher.

\begin{wraptable}[16]{r}{0.45\textwidth}
\centering
\vspace{-3em} \caption{
\textbf{Test Time Augmentation}. \textit{History} means cumulative of test samples. CLIP-B/16 (teacher).}
\label{tab:tda}
\setlength\tabcolsep{4pt} \resizebox{\linewidth}{!}{\begin{tabular}{l c c}
\toprule
\rowcolor{gray!20} 
Method & Requirements & ImageNet$\uparrow$ \\
\midrule
TDA{\scriptsize\textcolor{gray}{[CVPR'24]}}         & History & 69.5 \\
{DMN-ZS}{\scriptsize\textcolor{gray}{[CVPR'24]}}      & History & 72.2 \\
{DPE}{\scriptsize\textcolor{gray}{[NeurIPS'24]}}          & History  & 71.9 \\
{DynaPrompt}{\scriptsize\textcolor{gray}{[ICLR'25]}}          & History  & 72.9 \\
\midrule
{CoOp}{\scriptsize\textcolor{gray}{[IJCV'22]}} & Labeled Data & 71.5 \\
{CoCoOp}{\scriptsize\textcolor{gray}{[CVPR'22]}}       & Labeled Data  & 71.0 \\
{TPT + CoOp}  & Labeled Data  & 73.6 \\
{TPT + CoCoOp}& Labeled Data  & 71.0 \\
MTA + Coop{\scriptsize\textcolor{gray}{[CVPR'24]}}     & Labeled Data & 73.9 \\
\midrule
\rowcolor{black!5}    LQN   & History & 76.2 \\
\rowcolor{black!5}   LQN (Pretrained) & History & 77.5 \\
\bottomrule
\vspace{-4em}
\end{tabular}
}
\vspace{-5em} \end{wraptable}
These variants are optimized with Adam using a learning rate of $1 \times 10^{-4}$. Input images are normalized to ImageNet statistics. The total number of iterations is set to $T = 15$.  The GFLOPs are measured using Meta’s \textit{fvcore} package (FLOPs $\times 10^9$). 
To ensure statistical reliability, we report the mean accuracy of three runs with different seeds.

Given all possible spatial tokens ($n = H \times W \times L$), there are $\binom{n}{3} = O(n^3)$ plausible triplets $(src_1, src_2, dest)$ for recirculation. Iterating over all of these is computationally infeasible; instead we randomly sample $~5\%$ of these triplets, which provides sufficient performance gains, without incurring  excessive compute  costs (dropping tokens with minimal loss in performance~\citep{he2022masked}). 
All hyperparameters were chosen empirically.

\begin{table}[h]
\centering
\caption{\textbf{LQN architecture for TTT}: with input dimensions $h,w,c$ and feature dimension $d_{p}$: dimensionality of positional encoding. $s$: stride of convolutional filter in encoder, $d_{c}$: dimension of the intermediate token of teacher on which LQN learns. LQN contains two locations, $(src, dest)$, so additional $d_p$ term is added to the input. $\dagger:$ LQN-[src] contains only a single location, so this term is not added. }
\label{tab:architecture}
\resizebox{\linewidth}{!}{\begin{tabular}{l@{\hskip 15pt}l@{\hskip 15pt}l@{\hskip 15pt}l@{\hskip 15pt}l@{\hskip 15pt}l}
\toprule
& Layer &  Feature Dimension  & $n_{kernels}$ & Stride & Padding  \\ 
& &  \footnotesize{(H $\times$ W $\times$ C)} & & & \footnotesize{Input / Output}  \\
\midrule
& Input & $h$ $\times$ $w$ $\times$ $c$ \\
\midrule
Encoder     & Conv   & $h/s$ $\times$ $w/s$ $\times$ $d$ & 1& $s$ & 0 / 0  \\

\midrule
\multirow{8}{0.01\linewidth}{Decoder}    & Linear        &  $(h/s*w/s + d_p + d^{\dagger}_p)$ * $4096$  & - & - & -  \\ 
& Linear        &  $4096$ * $4096$  & - & - & -  \\ 
& Linear        &  $4096$ * $2048$  & - & - & -  \\ 
& Linear        &  $2048$ * $1024$  & - & - & -  \\ 
\midrule
Feature Projection Head     & Linear   & $1024$ * $d_{c}$ & - & - & -  \\
\midrule

\end{tabular}
}
\end{table}

\begin{table}[ht!]
\centering
\caption{\textbf{LQN hyperparameters}  during test-time-training.}
\label{tab:hparams}
\resizebox{\textwidth}{!}{\begin{tabular}{ll}
\toprule
Number of Test samples & 50000 (Imagenet Splits), variable for other datasets.\\
Testing iterations         & 15      \\
Batch Size              & 1        \\
Learning Rate         & 1e-4       \\
Optimizer         & Adam   
\\
\midrule
Feature Output size $d$         & $768/1024$       \\
Positional  Encoding size &$768/1024$\\
\midrule
Image/Crop Size         & 448        \\
Augmentations           & Normalization,  $\mu = (0.485, 0.456, 0.406)$, $\sigma= (0.229, 0.224,
0.225)$      \\

\midrule
Precision & fp16 (grad-scaled) \\
Num of Workers & 8 \\
Operating System & 1x rtx a6000 48GB/96GB ram/Ubuntu 22.04/2TB ssd/5TB HDD\\
\bottomrule
\end{tabular}
}
\label{tab:hyperparameter}
\end{table}

\begin{table}[ht!]
\centering
\caption{Values of $\alpha$, i.e. the weight of LCR loss for different datasets used in the experiments. $\alpha=0.7$, for the remaining datasets not shown here, for eg, COCO, ADE20k. Optimal values were found by grid search and choosing the parameters that gave lowest values of the LCR loss. Note that this procedure does not use any labels at test-time.  }
\label{tab:hparams_alpha}
\resizebox{\textwidth}{!}{\begin{tabular}{lccccccccccccc}
\toprule
Dataset & ImageNet & Flowers102 & DTD & Caltech101 & Aircraft & IN-A & IN-v2 & Pets & Sun397 & Eurosat & IN-R & UCF101 & Food101 \\
\midrule
$\alpha$   & 0.7 & 0.5 & 0.7 & 0.5 & 0.7 & 0.6 & 0.6 & 0.6 & 0.6 & 0.6 & 0.8 & 0.8 & 0.8 \\
$\epsilon$ (in LCR)  & 1e-7 & 1e-7 & 1e-7 & 1e-7 & 1e-7 & 1e-7 & 1e-7 & 1e-7 & 1e-7 & 1e-7 & 1e-7 & 1e-7 & 1e-7 \\
\bottomrule
\end{tabular}
}
\end{table}

\textbf{Hyperparameters:} All hyper-parameters utilized for LQN during test-time-training are detailed in \ref{tab:hparams}. We leveraged the seed 0/7/42 in most of our experiments. The weight matrices in LQN were initialized  from a normal distribution with $\mu=0$ and $\sigma= 0.01$. We tried initializing the net with other schemes such as Xavier initialization etc, but found normal-weight initialization to work the best. All our code has been written in Pytorch version 1.13.0. However, since we only use pytorch, and no external libraries, we expect that our codebase will also support more recent versions, for eg, PyTorch 2.0+. We will also release our implementation in Jax, in light of recent trends. We also note that performing test-time-training with $16$ bit floating point allows us to effectively use recent GPU architectures for eg, Ampere: they contain a larger number of tensor cores \textit{in addition} to CUDA cores which results in significant speedups during the experimentation process. Finally, we normalize an input image using standard Imagenet stats, and  \textit{don't resort to any other form of augmentation}, thereby making the pipeline far-simpler. 

Similarly, Table $\ref{tab:hparams_alpha}$ illustrates values of $\alpha$, i.e. the LCR loss for various datasets used in the experiments. Optimal values were found by grid search and choosing the parameters that gave lowest values of the LCR loss. Note that this procedure does not use any labels at test-time. We found that experiments are not particularly sensitive to datasets, and results lie within $0.2\%-0.5\%$ standard deviation if $\alpha=0.7$ is chosen, and experiments were run for different seeds.

\paragraph{ViT Encoder:} During our experiments in test-time-training, LQN relies on higher-dimensional intermediate token distilled from a teacher trained on a large-scale-dataset, often via contrastive image-text objectives. 
CLIP is a zero-shot model from OpenAI which contains a vision encoder, and a textual encoder. The textual encoder tokenizes input class names to features. Both image/text encoder project them to common dimensionality, and classification happens by measuring distances in contrastive space, thereby offering a higher degree of freedom, as opposed to training a class-sensitive linear-probe. It also offers the flexibility that the last layer of a neural net does not need to be re-trained to perform classification on a class never seen during training. This used to be a fundamental problem in earlier neural nets that applied softmax on a linear layer consisting of a fixed no of neurons. CLIP replaced this idea and instead treats classification as a problem of finding distances in the representational space. CLIP VIT-L outputs a CLS token of $768$ dimensions, while CLIP VIT-H outputs $1024$ dimensions.

\section{Details of the Datasets }
\label{sec:dataset_details}
Evaluating a model’s robustness to distribution shifts necessitates testing on datasets that feature a broad spectrum of perturbations, such as \textit{fog, snow, rain}, and other real-world variations. A common strategy is to apply synthetic corruptions to well-established test sets (e.g., ImageNet) to create benchmark splits suitable for controlled evaluation. Alternatively, new test sets may be manually compiled from online sources to capture modality changes, such as \textit{sketches} or \textit{artistic reinterpretations}. Below, we describe the key datasets used in this work to assess the robustness and generalization capabilities of LQN. These datasets consist of both classification and segmentation benchmarks.

\subsection{ Corruption \& Distribution Shift Benchmarks}
\label{sec:corruption_benchmark}
\noindent \textbf{CIFAR-10-C:} Comprising $10{,}000$ test samples from CIFAR-10, this benchmark introduces 15 corruption types (e.g., blur, noise, weather effects), each applied at 5 severity levels. Our evaluation focuses on the most difficult setting—level 5—due to computational constraints.

\noindent \textbf{ImageNet-C:} A widely used benchmark for corruption robustness based on the original ImageNet dataset. It includes 15 distortion types applied at 5 severity levels, affecting all 1,000 classes. The corruptions degrade image quality in ways that simulate real-world noise and artifacts.

\noindent \textbf{ImageNet-V2:} A re-collection of ImageNet-like samples from the web, preserving label distribution across 1,000 categories. The dataset consists of 10,000 images across three distinct splits, offering a testbed for evaluating generalization to naturally shifted data.

\noindent \textbf{ImageNet-A:} Contains 7,500 images from 200 categories that are hard for standard models like ResNet-50. These examples are “naturally adversarial”—real-world images that consistently cause misclassification.

\noindent \textbf{ImageNet-R:} Focuses on stylized renditions of ImageNet classes, including paintings, cartoons, and other artistic formats. It contains approximately 30,000 images across 200 categories and evaluates robustness to stylistic domain shifts.

\noindent \textbf{ImageNet-Sketch:} A challenging modality shift dataset containing ~50,000 black-and-white sketches corresponding to 1,000 ImageNet categories. It tests the model’s ability to recognize abstract shapes and contours, without relying on color or texture cues.

To ensure consistency with prior work (e.g., CLIP), we evaluate LQN using an ensemble of 80 handcrafted textual prompts across these ImageNet-derived benchmarks.

\subsection{Dense Prediction Benchmarks}
\label{sec:dense_prediction_benchmark}
\noindent \textbf{COCO (Common Objects in Context):} A large-scale dataset for object detection, segmentation, and captioning tasks. It features over 200,000 labeled images containing instances of 80 object categories in diverse, cluttered scenes. COCO is widely used to evaluate models’ ability to handle multi-object, real-world environments.

\noindent \textbf{Cityscapes:} Focused on urban street scenes, this dataset contains 5,000 finely annotated images from 50 different European cities. It includes 19 semantic classes relevant to autonomous driving (e.g., road, pedestrian, traffic light) and is primarily used for evaluating semantic segmentation under real-world conditions.

\noindent \textbf{ADE20K:} A challenging benchmark for semantic segmentation that includes 25,000 images annotated with over 150 object and stuff categories. The dataset spans indoor, outdoor, urban, and natural scenes, providing a diverse set of environments for evaluating dense prediction tasks.

\section{Additional Experiments }
\label{sec:additional_experiments}

\subsection{Variations of src in LQN}

\begin{wrapfigure}{r}{0.4\textwidth}
\centering
\vspace{-4em} \includegraphics[width=0.8\linewidth]{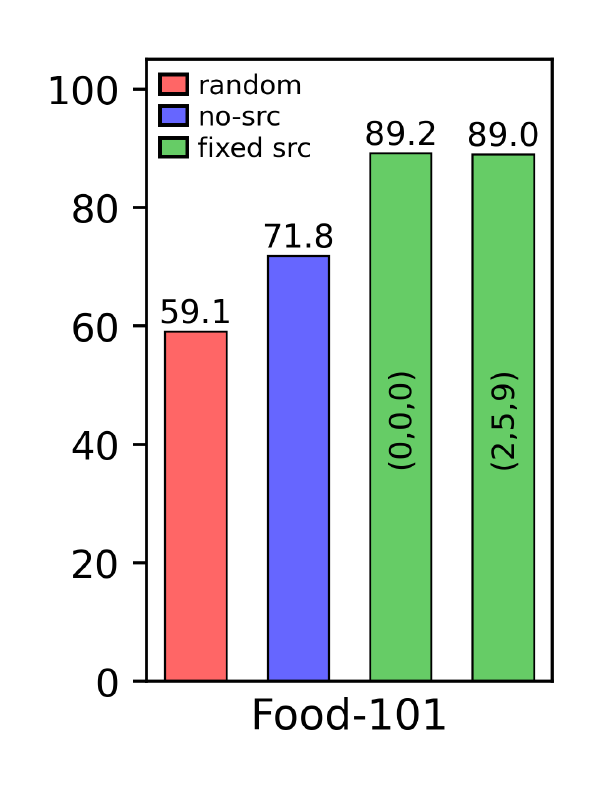}
\caption{Fixed src performs best.}
\label{fig:fixed_src}
\vspace{-4em} \end{wrapfigure}

During final iteration $N$, LQN decodes final layer features by setting src to (0,0,0). But why set a fixed source? In Fig\ref{fig:fixed_src}, we experiment with different values of $src$. Please note that this experiment was performed for Food101 classification, using CLIP VIT-B/16 teacher. Using a \textit{random} source merely achieves $59.1$, and not using the source at all only obtains $71.8$. Using a fixed source (0,0,0), (2,5,9) seems to obtain similar performance of $89.2$ and $89.0$ respectively. An intuitive reason may be that selecting a fixed source forces multiple destination coordinates (in the last layer) to be decoded \textit{relative} to an anchor.

\subsection{Concatenating embeddings performs better than  adding them to input} 

Recall that first MLP layer of LQN contains as input the test sample's representation $x'$, and positional encoding corresponding to $dest$. Instead of concatenating these inputs, we perform an additional experiment where positional encoding is added to x'. Specifically, given $x_{ood}$, we first convolve it with a CNN filter to yield $x'\in \mathds{R}^{h'\times w' \times 1}$. The resultant $x'$ is flattened and added to $p_{src}$, which similar to a transformer. Note that we project positional encoding through a learnt embedding layer, prior to adding to $x$. We don't decrease the dimension of encoding directly for the fairness of comparisons. We find that the TTT performance on ImageNet for LQN model using CLIP VIT-B/16 as a teacher \textit{drops} from $73.1$ to $62.8$. However, we do note that in the original VIT paper, \citep{dosovitskiy2021vit}, the best results were estimated by adding the positional encoding to the image features.

\subsection{Varying number of iterations $N$}

\begin{table}[H]
\centering
\caption{Model performance (Accuracy) across different iteration counts $N$.}
\begin{tabular}{lcccccc}
\toprule
\textbf{N} & 5 & 8 & 12 & 15 & 20 & 25 \\
\midrule
Accuracy & 80.5 & 83.9 & 88.6 & 89.2 & 87.8 & 81.0 \\
\bottomrule
\end{tabular}
\label{tab:n_iterations_performance}
\end{table}

On Food-101, we vary number of iterations $N$ while performing TTT on LQN, using CLIP VIT B/16 teacher. We find a peak performance of $89.2$ at $N=15$, with subsequent drops at other iterations. Please note that these are average scores on the entire dataset. 

\subsection{Ablations with Entropy Minimization}
\label{sec:ttt_ablations}

We use LQN-pretrained variant, and evaluate on ImageNet val split with a CLIP VIT-B/16 teacher. An entropy-minimization objective (similar to TPT) on LQN \textit{drops} performance from $74.5$ to $67.4$, indicating that Cosine might serve as a better loss. 

\begin{figure*}[ht!]
\centering
\includegraphics[width=0.6\textwidth]{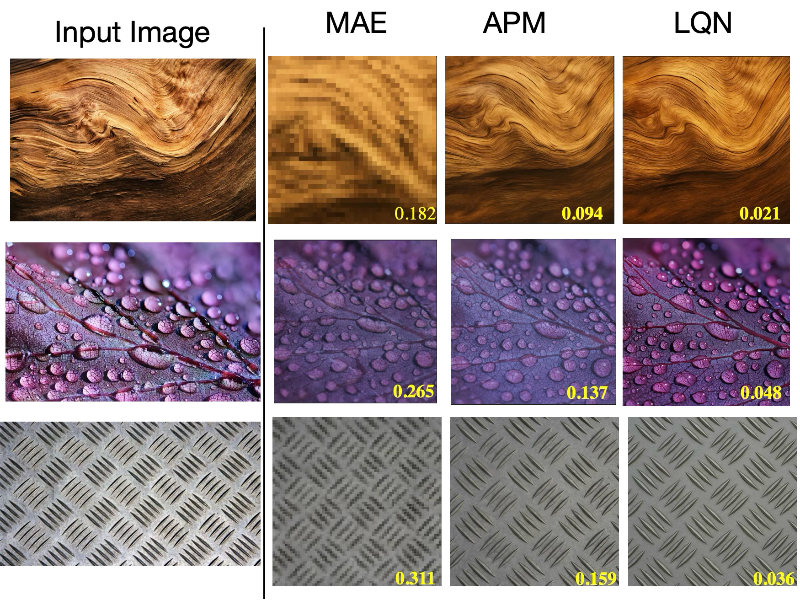}
\caption{\textbf{Qualitative results on LQN.} LQN reconstructs RGB images with \textit{lowest} pixelwise loss (numbers marked in yellow) as compared to MAE/APM. This illustrates the extreme scenario when input images contain repeating textures and tests whether the model encodes semantics consistently. Note that these results are \textit{not} cherry-picked.}
\label{fig:repeating_texture}
\end{figure*}

\subsection{Evaluating semantic-consistency on repeated textures}
\label{sec:semantic_consistency_texture}
LQN has the ability to process different locations of an image in parallel. It consists of two procedures aka PAD, and LCR. During the PAD phase, different locations `never' interact among themselves other than mere weight-sharing of LQN across different positions. This raises the question: How can the network learn long-range dependencies, and how does it actually perform in cases of repeated textures? .  Indeed, repeated textures have presented challenges to various computer-vision systems \citep{wang2025vggt}, where the neural nets are unable to find ``correspondences" between regions that have repeating textures. To investigate this, we make following changes to LQN: 1)  we \textit{remove} the LCR procedure, and only test the PAD procedure. 2) LQN is made to perform RGB reconstruction similar to MAE. We train LQN on ImageNet val, and test the rgb-reconstruction performance on a manually curated test-set of images with repeated textures. Fig \ref{fig:repeating_texture} shows that LQN incurs lowest $L_2$ error during reconstruction.

\subsection{Evaluating Latency on Constrained-Devices}

In Tab\ref{tab:edge_device}, we report GFlops/ Memory occupied/ time taken for single forward pass, as well as total time for TTT, when running LQN using EoMT teacher for panoptic segmentation on COCO. We find that around $66.38\%$
is consumed through first forward pass through the teacher (at N = 0), and inference of the segmentation mask (at N = 15). For an alternate classification task (like on imagenet val), the inference happens in latentspace, and \textit{does not} require any forward pass through a separate segmentation head, which is why it can be done quickly.

\begin{table}[H]
\centering
\caption{Memory and GFlop Analysis.}
\begin{tabular}{lccccc}
\toprule
& \textbf{Memory} & \textbf{ORIN} & \textbf{AGX Thor} & \textbf{Ampere} & \textbf{GFlops} \\
\midrule
EoMT (teacher) & 2.1 GB & 0.5 sec & 0.1sec & 0.08sec & 4146 \\
LQN (student) / iter & 600 MB & 0.68sec & 0.37sec & 0.29sec & 24.5 \\
LQN total (15 iters) & 2.7 GB & 8.9sec & 4.3sec & 2.88sec & 4384 \\
\bottomrule
\end{tabular}
\label{tab:edge_device}
\end{table}

\subsection{Applicability to different domains}

Table \ref{tab:medical_di} evaluates the applicability of LQN to a different domain that the teacher has not seen. For eg, here we experiment with a medical dataset (Kaggle Pneumothorax Lung Dataset), consisting of binary classification of chest x-rays for collapsed lungs vs healthy lungs. Zero-shot accuracy of CLIP VIT-L is 74.9. By distilling different number of layers, LQN is able to improve the teacher network to $77.3\%$. Please note that, in theory, it would be possible to use a teacher pre-trained explicitly on medical data (eg, MedCLIP). However, that violates the test for applicability to different domains.

\begin{table}[H]
\centering
\caption{Accuracy comparison across different layer configurations.} \begin{tabular}{lccccc}
\toprule
\textbf{Model} & \textbf{CLIP ViT-L} & \textbf{2 Layers} & \textbf{4 Layers} & \textbf{8 Layers} & \textbf{10 Layers} \\
\midrule
Accuracy & 74.9 & 75.8 & 76.5 & 77.3 & 77.0 \\
\bottomrule
\end{tabular}
\label{tab:medical_di} \end{table}

\subsection{Applicability to Non-VIT backbones}

In Tab\ref{tab:vision_mamba_results}, we evaluate applicability of LQN to distill from Non-VIT backbones. We benchmark Vision Mamba on ImageNet val set. LQN with Vision-Mamba as a teacher is able to improve performance from $78.3 \rightarrow 80.9$ for \textit{tiny variant}. Similarly, with Vision-Mamba \textit{Small} variant, we see performance improve  from $81.4 \rightarrow 83.2$.

\begin{table}[ht!]
\centering
\caption{Performance comparison of Vision-Mamba models with LQN.}
\begin{tabular}{lcc}
\toprule
\textbf{Model} & \textbf{Base} & \textbf{+ LQN} \\
\midrule
Vision-Mamba-T & 78.3 & 80.9 \\
Vision-Mamba-S & 81.4 & 83.2 \\
\bottomrule
\end{tabular}
\label{tab:vision_mamba_results}
\end{table}

\subsection{Experiments  on ImageNet-C}
\label{sec:imagenets_other_levels}
Following TTT-MAE\citep{ttt_mae}, we evaluate our method on ImageNet-C. ImageNet-C is a dataset which consists of $15$ types of corruptions applied to the original ImageNet validation  set. As evidenced in Tab \ref{tab:imagenet_c}, we obtain the highest accuracy of $53.0$ on the highest severity level $5$, thereby showcasing the efficacy of LQN.

\begin{table*}[ht!]
\caption{\textbf{LQN's performance on ImageNet-C, level 5}. The first three rows are fixed models without test-time training. The third row, ViT probing, is the baseline used in \citep{ttt_mae}. A \ding{51} in P means that method leveraged \textbf{pre-trained weights} on clean variant of train set aka, Image-net and downstream-ttt on corrupted version. OpenCLIP VIT-L/14 is generally more robust. LQN does better than APM.}
\label{tab:imagenet_c}
\centering    
\resizebox{\linewidth}{!}{\begin{tabular}{l*{18}{c}}
\toprule
& P & brigh & cont & defoc & elast & fog & frost & gauss & glass & impul & jpeg & motn & pixel & shot & snow & zoom & Average \\
\midrule
\rowcolor[HTML]{EFEFEF}
Joint Train & \ding{51} & 62.3 & 4.5 & 26.7 & 39.9 & 25.7 & 30.0 & 5.8 & 16.3 & 5.8 & 45.3 & 30.9 & 45.9 & 7.1 & 25.1 & 31.8 & 24.8 \\
\rowcolor[HTML]{EFEFEF}
Fine-Tune & \ding{51} & 67.5 & 7.8 & 33.9 & 32.4 & 36.4 & 38.2 & 22.0 & 15.7 & 23.9 & 51.2 & 37.4 & 51.9 & 23.7 & 37.6 & 37.1 & 33.7 \\
\rowcolor[HTML]{EFEFEF}
ViT Probe & \ding{51} & 68.3 & 6.4 & 24.2 & 31.6 & 38.6 & 38.4 & 17.4 & 18.4 & 18.2 & 51.2 & 32.2 & 49.7 & 18.2 & 35.9 & 32.2 & 29.2 \\
\rowcolor[HTML]{EFEFEF}
TTT-MAE & \ding{51} & 69.1 & 9.8 & 34.4 & 50.7 & 44.7 & 50.7 & 30.5 & 36.9 & 32.4 & 63.0 & 41.9 & 63.0 & 33.0 & 42.8 & 45.9 & 44.4 \\
\midrule 
OpenCLIP VIT-L/14(t) & \ding{55} & 71.9 & 47.0 & 50.3  & 32.7 & 58.3 & 46.9 & 26.0 & 26.5 & 28.1 & 62.7 & 37.7 & 58.3 & 28.2 & 50.4 & 37.9 & 42.1 \\
APM & \ding{55} & 77.4 & 51.9 & 56.6 & 37.9 & 64.8 & 53.2 & 28.7 & 31.4 & 33.0 & 68.4 & 44.1 & 64.5 & 33.1 & 56.9 & 43.9 & 50.3 \\
\hline
LQN (Ours)  & \ding{55} & \textbf{80.1} & \textbf{55.2} & \textbf{60.4} & \textbf{41.5} & \textbf{68.7} & \textbf{56.4} & \textbf{31.6} & \textbf{34.2} & \textbf{36.0} & \textbf{71.9} & \textbf{47.2} & \textbf{67.9} & \textbf{36.3} & \textbf{60.3} & \textbf{47.1} & \textbf{53.0} \\
\bottomrule
\end{tabular}
}
\end{table*}

\begin{table*}[ht!]
\caption{\textbf{Performance on ImageNet-C, level 4}. The first two rows are from the supplementary materials of \citep{ttt_mae}. A \ding{51} in column P indicates use of \textbf{pre-trained weights} on clean ImageNet followed by TTT on corrupted inputs. OpenCLIP ViT-L/14 shows stronger robustness than earlier models. LQN does better than APM.}
\label{tab:imagenet_c_level_4}
\centering    
\resizebox{\linewidth}{!}{\begin{tabular}{l*{18}{c}}
\toprule
& P & brigh & cont & defoc & elast & fog & frost & gauss & glass & impul & jpeg & motn & pixel & shot & snow & zoom & Average \\
\midrule
\rowcolor[HTML]{EFEFEF}
Baseline & \ding{51} & 73.1 & 33.1 & 35.8 & 56.9 & 54.2 & 45.2 & 39.6 & 26.0 & 38.2 & 62.0 & 43.2 & 60.3 & 32.2 & 44.2 & 40.7 & 47.4 \\
\rowcolor[HTML]{EFEFEF}
TTT-MAE & \ding{51} & 72.7 & 39.6 & 45.7 & 64.9 & 58.3 & 52.6 & 48.5 & 42.8 & 47.6 & 67.0 & 50.5 & 66.6 & 42.4 & 45.7 & 51.5 & 53.2 \\
\midrule 
OpenCLIP VIT-L/14 & \ding{55} & 74.2 & 64.2 & 58.7 & 57.8 & 66.3 & 52.8 & 45.3 & 34.6 & 45.2 & 68.9 & 46.6 & 63.9 & 41.1 & 56.2 & 45.6 & 54.8 \\
APM               & \ding{55} & 79.2 & 70.4 & 64.9 & 63.7 & 72.3 & 58.6 & 51.2 & 40.4 & 51.3 & 74.1 & 53.0 & 70.0 & 46.7 & 62.5 & 51.8 & 59.6 \\
\hline
LQN (Ours)   & \ding{55} & \textbf{80.9} & \textbf{72.1} & \textbf{66.5} & \textbf{65.2} & \textbf{73.9} & \textbf{59.9} & \textbf{52.9} & \textbf{41.9} & \textbf{52.8} & \textbf{75.3} & \textbf{54.2} & \textbf{71.4} & \textbf{47.9} & \textbf{63.9} & \textbf{53.3} & \textbf{61.2} \\
\bottomrule
\end{tabular}
}
\end{table*}

\begin{table*}[ht!]
\caption{\textbf{Performance on ImageNet-C, level 3}. The first two rows are from the supplementary materials of \citep{ttt_mae}. A \ding{51} in column P indicates that the method used \textbf{pre-trained weights} on clean ImageNet and applied TTT on the corrupted set. OpenCLIP ViT-L/14 is more robust than earlier models. LQN does better than APM.}
\label{tab:imagenet_c_level_3}
\centering    
\resizebox{\linewidth}{!}{\begin{tabular}{l*{18}{c}}
\toprule
& P & brigh & cont & defoc & elast & fog & frost & gauss & glass & impul & jpeg & motn & pixel & shot & snow & zoom & Average \\
\midrule
\rowcolor[HTML]{EFEFEF}
Baseline & \ding{51} & 75.8 & 62.7 & 49.5 & 67.1 & 59.8 & 47.6 & 57.1 & 35.0 & 57.4 & 68.6 & 60.2 & 70.1 & 54.3 & 54.7 & 48.0 & 57.6 \\
\rowcolor[HTML]{EFEFEF}
TTT-MAE & \ding{51} & 75.8 & 64.4 & 59.4 & 71.2 & 64.0 & 54.0 & 63.6 & 50.7 & 64.2 & 71.3 & 64.2 & 73.1 & 61.8 & 58.0 & 57.4 & 64.4 \\
\midrule 
OpenCLIP VIT-L/14 & \ding{55} & 75.8 & 71.8 & 65.5 & 67.7 & 69.0 & 54.7 & 58.9 & 42.4 & 59.5 & 72.8 & 59.9 & 69.7 & 58.2 & 63.5 & 51.8 & 62.5 \\
APM               & \ding{55} & 80.5 & 77.2 & 71.3 & 73.3 & 74.8 & 60.6 & 64.7 & 48.5 & 65.4 & 77.8 & 61.6 & 75.2 & 64.1 & 69.3 & 58.0 & 68.5 \\
\hline
LQN (Ours)   & \ding{55} & \textbf{81.9} & \textbf{78.8} & \textbf{72.8} & \textbf{74.9} & \textbf{76.2} & \textbf{61.9} & \textbf{66.3} & \textbf{49.9} & \textbf{66.9} & \textbf{79.1} & \textbf{62.8} & \textbf{76.7} & \textbf{65.5} & \textbf{70.5} & \textbf{59.5} & \textbf{69.9} \\
\bottomrule
\end{tabular}
}
\end{table*}

\begin{table*}[ht!]
\caption{\textbf{Performance on ImageNet-C, level 2}. The first two rows are from the supplementary materials of \citep{ttt_mae}. A \ding{51} in column P indicates that the method used \textbf{pre-trained weights} on clean ImageNet and performed TTT on the corrupted version. OpenCLIP ViT-L/14 is generally more robust than earlier models. LQN does better than APM. }
\label{tab:imagenet_c_level_2}
\centering    
\resizebox{\linewidth}{!}{\begin{tabular}{l*{18}{c}}
\toprule
& P & brigh & cont & defoc & elast & fog & frost & gauss & glass & impul & jpeg & motn & pixel & shot & snow & zoom & Average \\
\midrule
\rowcolor[HTML]{EFEFEF}
Baseline & \ding{51} & 77.4 & 71.2 & 62.3 & 51.0 & 66.3 & 58.4 & 68.6 & 59.2 & 64.9 & 70.4 & 70.6 & 74.7 & 66.2 & 54.2 & 55.2 & 64.1 \\
\rowcolor[HTML]{EFEFEF}
TTT-MAE & \ding{51} & 77.8 & 71.5 & 69.4 & 49.7 & 69.8 & 62.7 & 72.5 & 66.4 & 70.0 & 72.7 & 72.3 & 76.2 & 70.6 & 58.7 & 63.6 & 68.3 \\
\midrule 
OpenCLIP VIT-L/14 & \ding{55} & 76.6 & 74.4 & 71.4 & 53.8 & 72.0 & 62.6 & 67.6 & 64.0 & 64.6 & 73.8 & 69.0 & 72.8 & 66.4 & 61.8 & 58.3 & 66.1 \\
APM               & \ding{55} & 81.1 & 79.4 & 76.6 & 59.4 & 77.3 & 68.2 & 73.1 & 70.0 & 70.3 & 78.6 & 74.5 & 77.8 & 72.0 & 67.8 & 64.3 & 72.4 \\
\hline
LQN (Ours)   & \ding{55} & \textbf{82.4} & \textbf{81.0} & \textbf{78.4} & \textbf{60.9} & \textbf{78.9} & \textbf{69.8} & \textbf{74.7} & \textbf{71.3} & \textbf{71.7} & \textbf{80.1} & \textbf{75.8} & \textbf{79.2} & \textbf{73.4} & \textbf{69.1} & \textbf{65.8} & \textbf{74.1} \\
\bottomrule
\end{tabular}
}
\end{table*}

\begin{table*}[ht!]
\caption{\textbf{APM and LQN performance on ImageNet-C, level 1}. The first two rows are reproduced from the supplementary materials of \citep{ttt_mae}. A \ding{51} in column P indicates that the method used \textbf{pre-trained weights} on clean ImageNet and applied TTT on the corrupted set. OpenCLIP VIT-L/14 is generally more robust than earlier models. LQN does better than APM.}
\label{tab:imagenet_c_level_1}
\centering    
\resizebox{\linewidth}{!}{\begin{tabular}{l*{18}{c}}
\toprule
& P & brigh & cont & defoc & elast & fog & frost & gauss & glass & impul & jpeg & motn & pixel & shot & snow & zoom & Average \\
\midrule
\rowcolor[HTML]{EFEFEF}
Baseline & \ding{51} & 78.5 & 74.5 & 68.1 & 73.9 & 70.5 & 70.6 & 74.8 & 68.6 & 72.3 & 73.0 & 75.2 & 75.9 & 73.6 & 69.3 & 63.7 & 71.4 \\
\rowcolor[HTML]{EFEFEF}
TTT-MAE & \ding{51} & 78.9 & 74.7 & 72.5 & 74.7 & 72.9 & 72.2 & 76.8 & 72.2 & 75.5 & 74.5 & 75.8 & 77.0 & 75.9 & 71.9 & 69.3 & 73.1 \\
\midrule 
OpenCLIP VIT-L/14 & \ding{55} & 77.3 & 75.4 & 73.5 & 73.1 & 73.5 & 71.4 & 71.9 & 70.2 & 69.9 & 75.1 & 73.7 & 74.2 & 71.9 & 71.2 & 65.2 & 71.1 \\
APM               & \ding{55} & 81.6 & 80.3 & 78.6 & 78.0 & 78.6 & 76.6 & 77.2 & 75.7 & 75.1 & 79.6 & 78.7 & 79.1 & 76.9 & 76.4 & 70.7 & 76.0 \\
\hline
LQN (Ours)   & \ding{55} & \textbf{83.2} & \textbf{82.0} & \textbf{80.3} & \textbf{79.6} & \textbf{80.1} & 78.0 & \textbf{78.6} & \textbf{77.0} & \textbf{76.5} & \textbf{80.9} & \textbf{80.2} & \textbf{80.5} & \textbf{78.4} & \textbf{77.9} & 72.3 & \textbf{77.6} \\
\bottomrule
\end{tabular}
}
\end{table*}

\subsection{Experiments on CIFAR10-C}
\label{sec:cifar10c}
We report more results on CIFAR-10-C in Table \ref{tab:cifar_10_c}, where LQN model gets the lowest error rate of $13.0$.
\begin{table*}[ht!]
\caption{\textbf{CIFAR-10-C} results at \textit{highest} severity level of 5. We report Error Rate (\%, lower is better). The t- model acts as the teacher for APM and LQN variants. TTT was performed on the test set with randomly initialized weights. APM and LQN weights were reinitialized after each TTT iteration to prevent information leakage. }
\label{tab:cifar_10_c}
\centering    
\resizebox{\textwidth}{!}{\begin{tabular}{l|c|c|c|c|c|c|c|c|c|c|c|c|c|c|c|c|c}
\toprule
Method     & orig & gauss & shot & impul & defoc & glass & motn & zoom & snow & frost & fog  & brit & contr & elas & pixel & jpeg & Avg \\
\hline 
TTT-Online & 8.2  & 25.8  & 22.6 & 30.6  & 14.6  & 34.4  & 18.3 & 17.1 & 20.0 & 18.0  & 16.9 & 11.2 & 15.6  & 21.6 & 18.1  & 21.2 & 19.1 \\
UDA-SS     & 9.0  & 28.2  & 26.5 & 20.8  & 15.6  & 43.7  & 24.5 & 23.8 & 25.0 & 24.9  & 17.2 & 12.7 & 11.6  & 22.1 & 20.3  & 22.6 & 21.4 \\
\hline
Zeroshot & & & & & & & & & & & & & & & & & \\
CLIP ViT-L/14     & 4.63  & 35.4  & 32.3 & 21.9  & 19.3  & 49.7  & 19.3 & 17.3  & 17.0 & 15.1  & 21.6 & 8.4 & 15.9  & 34.6 & 25.0  & 27.4  & 24.5 \\
\hline 
CLIP ViT-L/14 (t) & & & & & & & & & & & & & & & & & \\
APM               & 3.5 & 21.9 & 30.1 & 13.7 & 15.2 & 34.1 & 11.9 & 11.1 & 15.0 & 9.0 & 13.5 & 5.8 & 9.5  & 23.0 & 15.8 & 17.0 & 14.8 \\
\hline
LQN (Ours)   & 2.9  & 18.8  & 20.0 & 12.5  & 14.1  & 30.2  & 10.5 & 10.1  & 13.4 & 8.1  & 11.7 & 4.9 & 8.1  & 19.3 & 13.6  & 14.3 & 13.0 \\
\bottomrule
\end{tabular}
}
\end{table*}

\clearpage
\section{Visualization of Segmentation Masks }
\label{sec:qualitative_demonstration}

\begin{figure*}[ht]
\centering
\includegraphics[width=0.95\textwidth]{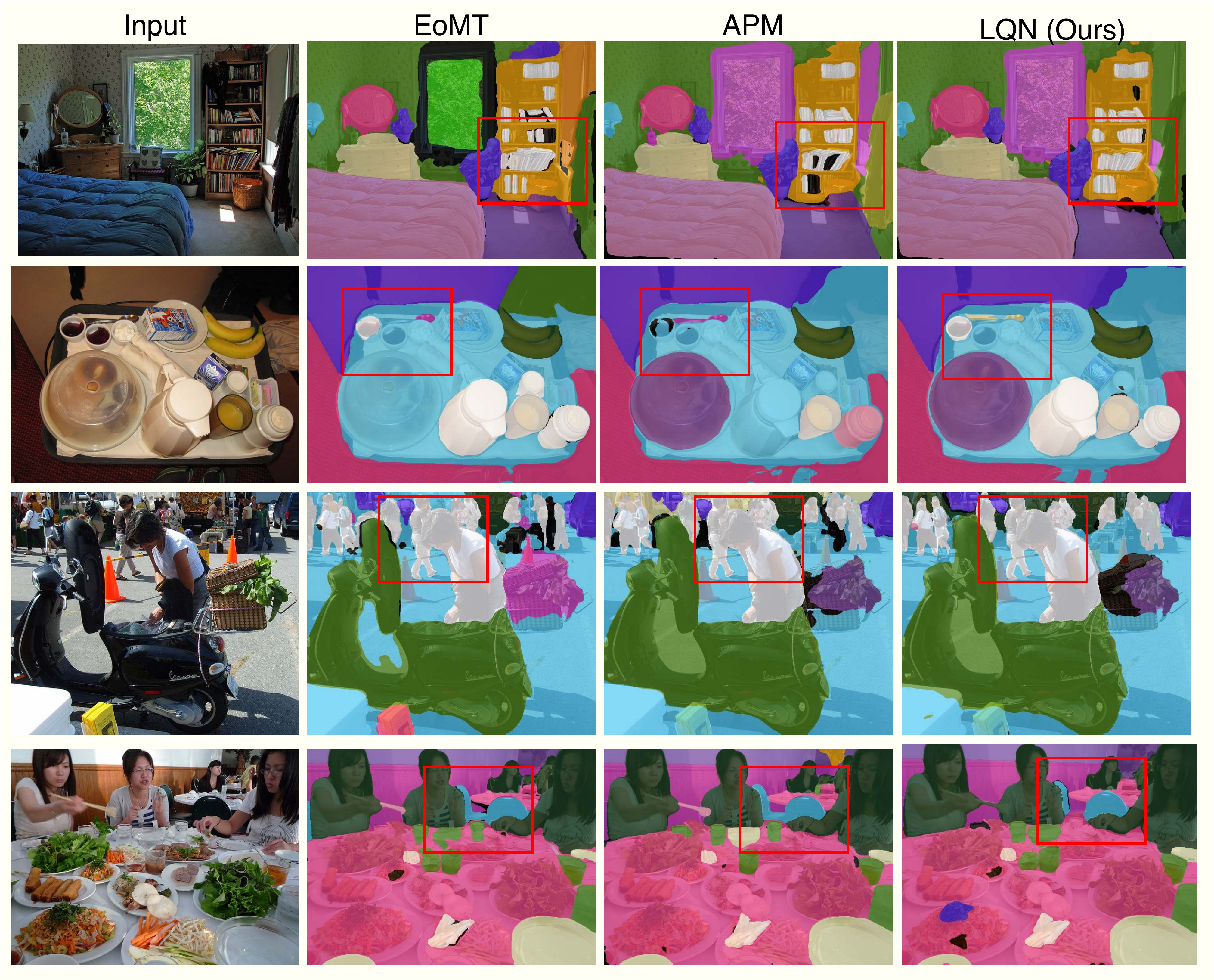}
\caption{\textbf{Qualitative results on LQN.} Panoptic segmentation on COCO-Val set. LQN obtains more semantically-detailed masks than the EoMT/APM baselines via test-time-training. Masks visualized after 15 iterations of TTT on both APM/LQN. EoMT is a fully-supervised, \textit{fixed} baseline. TTT on APM/LQN is performed with EoMT as the teacher. The red regions are the regions where a kind reader can focus on to see the comparison in the prediction quality. Starting from above, LQN easily segments the books in the bookshelf as white region, spoon as a white outline on the plate, distinguishes between the people in the background, and easily partitions chairs into two distinct parts, whereas other EoMT and APM group them together. }
\label{fig:qualitative-lqn-1}
\end{figure*}

\begin{figure*}[ht]
\centering
\includegraphics[width=0.95\textwidth]{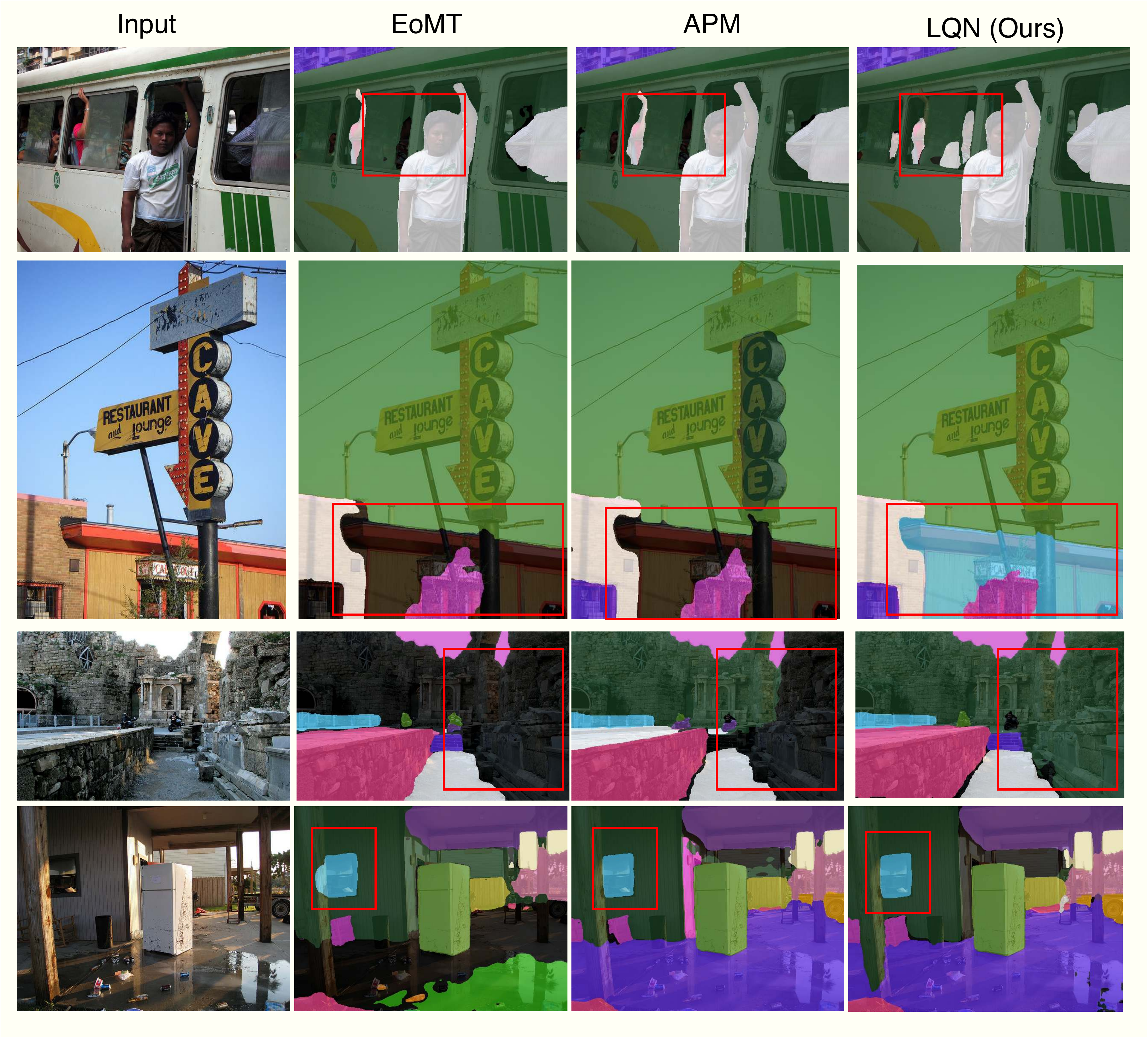}
\caption{\textbf{Qualitative results on LQN.} Panoptic segmentation on COCO-Val set. LQN obtains more semantically detailed masks than the EoMT/APM baselines via test-time training. Masks visualized after 15 iterations of TTT on both APM/LQN. EoMT is a fully-supervised, \textit{fixed} baseline. TTT on APM/LQN is performed with EoMT as the teacher. The red regions are the regions where a kind reader can focus on to see the comparison in the prediction quality. Starting from above, (first row) LQN gets the passengers in the bus in the background, even though they are partially occluded; (second row) easily segments the building into the blue regions; (third row) partitions buildings into a green colored region, whereas the other two methods don't detect it as all/term it as background; (fourth row) gets the window in the proper shape. }
\label{fig:qualitative-lqn-2}
\end{figure*}

\begin{figure*}[ht]
\centering
\includegraphics[width=0.95\textwidth]{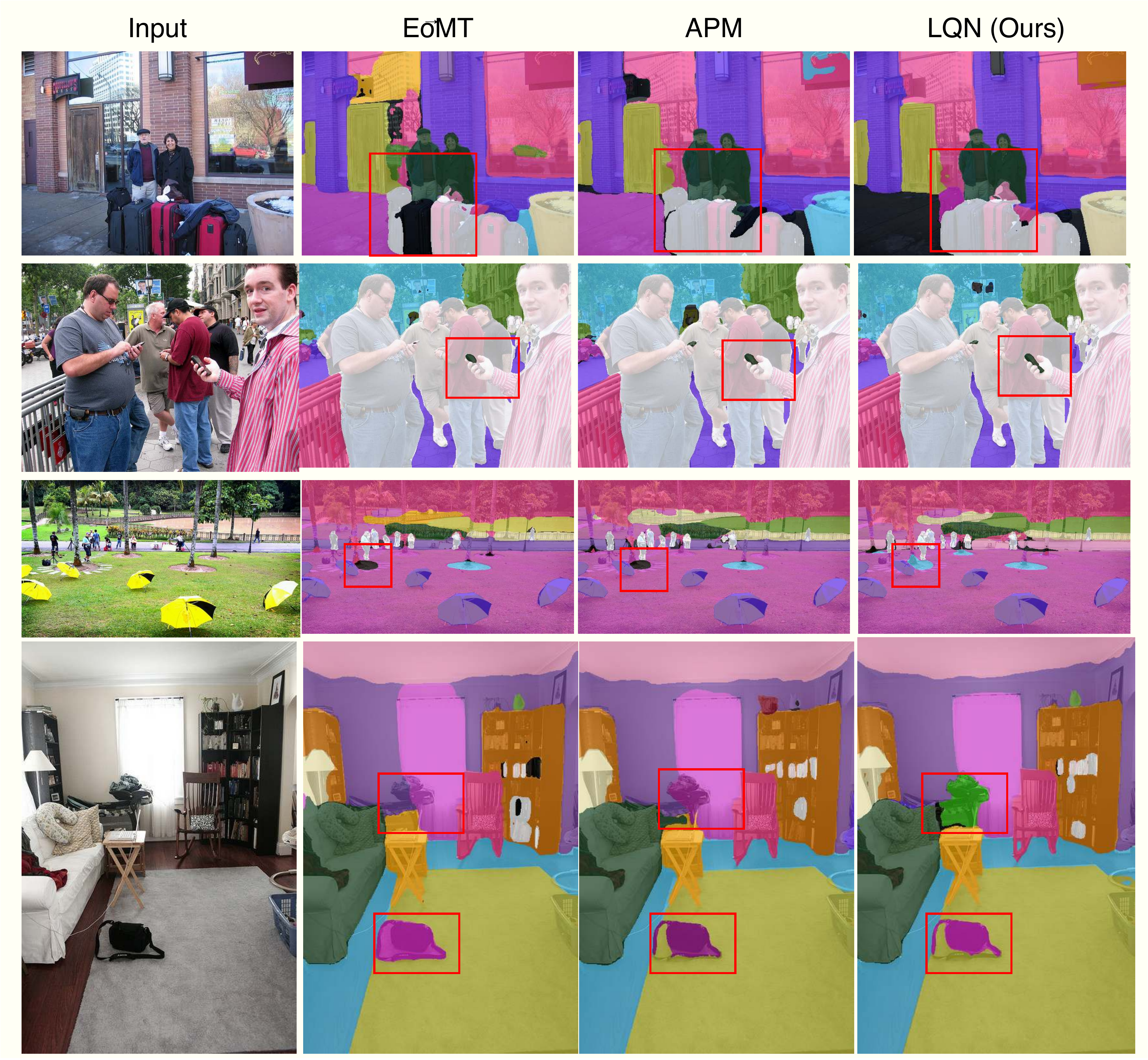}
\caption{\textbf{Qualitative results on LQN.} Panoptic segmentation on COCO-Val set. LQN obtains more semantically-detailed masks than the EoMT/APM baselines via test-time-training. Masks visualized after 15 iterations of TTT on both APM/LQN. EoMT is a fully-supervised, \textit{fixed} baseline. TTT on APM/LQN is performed with EoMT as the teacher. The red regions are the regions where a kind reader can focus on to see the comparison in the prediction quality. Starting from above, (first row) LQN properly segments the suitcases, (second row) Precise boundaries of the cellphone in the person's hand, even though part of the cellphone is gripped very tightly by the person's hand.  (third row) segments the base of the tree, whereas the other two methods don't detect it at all (fourth row) is able to understand the fine regions which correspond to the bag/floor. In contrast, EoMT/APM confuse that some part of the purse is also the background. }
\label{fig:qualitative-lqn-3}
\end{figure*}

\end{document}